\documentclass[11pt]{article}

\usepackage[final]{acl}

\usepackage{times}
\usepackage{latexsym}
\usepackage{amsmath}
\usepackage{amssymb}
\usepackage{booktabs}
\usepackage{multirow}
\usepackage{array}

\usepackage{pifont}       
\newcommand{\xmark}{\ding{55}} 

\usepackage{mdframed}

\usepackage{algorithm}
\usepackage{algpseudocode}
\algrenewcommand\algorithmiccomment[1]{\hfill{\footnotesize$\triangleright$~#1}}

\algrenewcommand\algorithmicindent{1.2em}

\usepackage[T1]{fontenc}

\usepackage[utf8]{inputenc}

\usepackage{microtype}

\usepackage{inconsolata}

\usepackage{graphicx}
\usepackage{multirow}
\usepackage{subcaption}

\title{ReEfBench: Quantifying the Reasoning Efficiency of LLMs}

\author{Zhizhang Fu $^*$ \\
  Westlake University \\
  \texttt{fuzhzihang@westlake.edu.cn} \\
  \And
  Yuancheng Gu $^*$ \\
  Imperial College London \\
  \texttt{yuancheng.gu22@}\\ \texttt{alumni.imperial.ac.uk} \\
  \And
  Chenkai Hu \\
  New York University \\
  \texttt{ckh326@nyu.edu} \\
  \AND
  Hanmeng Liu \\
  Hainan University \\
  \texttt{liuhanmeng@hainanu.edu.cn} \\
  \And
  Yue Zhang † \\
  Westlake University \\
  \texttt{zhangyue@westlake.edu.cn} \\
  }

\begin{document}
\maketitle

\def\thefootnote{*}\footnotetext{These authors contributed equally to this work}\def\thefootnote{\arabic{footnote}}

\def\thefootnote{†}\footnotetext{Corresponding author}\def\thefootnote{\arabic{footnote}}

\begin{abstract}
Test-time scaling has enabled Large Language Models (LLMs) to tackle complex reasoning, yet the limitations of current Chain-of-Thought (CoT) evaluation obscures whether performance gains stem from genuine reasoning or mere verbosity. To address this, (1) we propose a novel neuro-symbolic framework for the non-intrusive, comprehensive process-centric evaluation of reasoning. (2) Through this lens, we identify four distinct behavioral prototypes and diagnose the failure modes. (3) We examine the impact of inference mode, training strategy, and model scale. Our analysis reveals that extended token generation is not a prerequisite for deep reasoning. Furthermore, we reveal critical constraints: mixing long and short CoT data in training risks in premature saturation and collapse, while distillation into smaller models captures behavioral length but fails to replicate logical efficacy due to intrinsic capacity limits.

\end{abstract}

\begin{figure*}[t]
    \centering
    \includegraphics[width=0.93\textwidth]{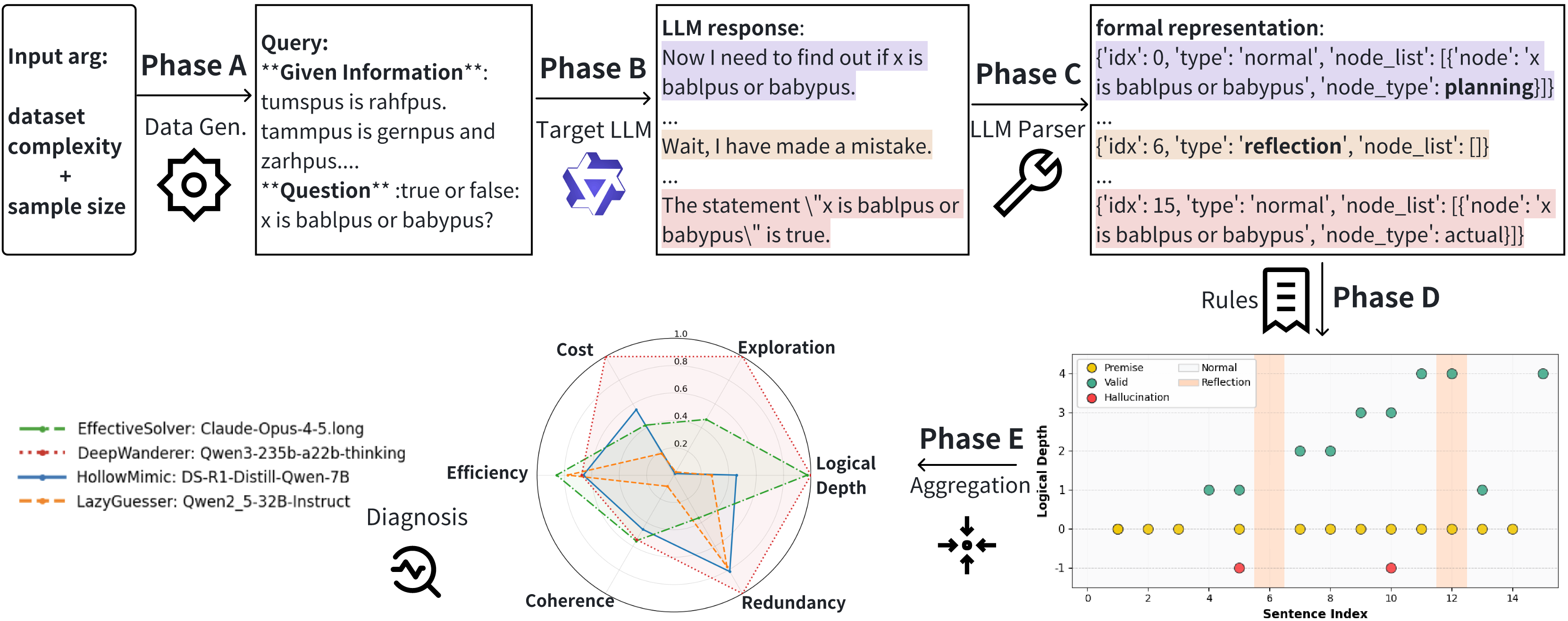}
    \caption{Overview of our framework. We generate scalable, controllable first-order logic (FOL) data that enables precise verification of logical depth (Phase A). Target LLM generates a response for each query (Phase B). The pipeline parses a target LLM’s response into formal representations (Phase C), computes its logical depth and correctness via rule-based verifiers (Phase D), evaluates the normalized output along six dimensions—Logical Depth, Cost, Exploration, Efficiency, Coherence, and Redundancy—and finally classifies it into one of four behavioral prototypes: EffectiveSolver, DeepWanderer, HollowMimic, and LazyGuesser (Phase E).}
    \label{fig:head_fig}
\end{figure*}

\section{Introduction}
Test-time scaling~\citep{zhang2025survey} has empowered LLMs to tackle complex problems by allocating more compute to reasoning steps~\citep{openai2024learning}. However, this paradigm reveals a perplexing inefficiency: models often exhibit ``overthinking'', generating protracted thinking chains even for trivial tasks like calculating $2 + 3$~\citep{chen2025do}. 
Given this observation, however, further \textbf{quantifying} the discrepancy systematically remains challenging due to the lack of process-centric evaluation, which is necessary for distinguishing genuine cognitive depth from mere computational inflation.

To this end, a simple intuition is to measure efficiency using $W = P \cdot t$, where $t$ represents the computational consumption (e.g., token or step count) and $W$ denotes the abstract ``Logical Depth'' achieved. Consequently, $P$ represents the \textbf{reasoning efficiency}—the logical gain per unit of computation. Efficient models maximize $P$, achieving the necessary logical depth $W$ with minimal cost $t$, whereas ``overthinking'' models increase $t$ without a proportional gain in $W$. 

While it is relatively easier to quantify the Cost ($t$) (e.g., token or step count), accurately quantifying Logical Depth ($W$) requires a specialized evaluation framework with three key attributes: (1) \textbf{A Formal Basis} (e.g., First-Order Logic), to ensure the reasoning path has an objective, calculable logical depth; (2) \textbf{Decoupling of Logic and Knowledge}, to evaluate pure reasoning capabilities without interference from knowledge; and (3) \textbf{Controllability and Scalability}, to generate problems of precise and varying levels of logical depth. To this end, existing datasets such as ProntoQA~\citep{saparov2023language} quantify reasoning over FOL, but suffer from limitations in scalability and, crucially, lack an evaluation mechanism to calculate logical depth and extensive process behaviors. 


To address this, we construct a comprehensive neuro-symbolic evaluation framework, Reasoning Efficiency Bench (ReEfBench), as illustrated in Figure~\ref{fig:head_fig}. Starting with the generation of test instances (Phase A) and the acquisition of model responses (Phase B), our pipeline utilizes an LLM parser to decompose the reasoning text into logical nodes (Phase C). We then apply deterministic rules to identify node-level attributes such as logical depth (Phase D), which are finally aggregated to compute comprehensive \textit{Behavioral Indicators} (Phase E). This approach combines the generalization of neural models with the rigor of symbolic logic, enabling a non-intrusive and quantifiable evaluation of the reasoning process.



We use this framework to conduct a comprehensive evaluation of 25 open-source and closed-source models, quantifying their reasoning depth and corresponding cost. Given the quantitative results, we identify four behavioral prototypes: the \textbf{Deep Wanderer} (high token consumption, exhaustive exploration; e.g., Qwen3-235B-thinking), the \textbf{Effective Solver} (high efficiency, precise reasoning; e.g., Claude-Opus-4.5), the \textbf{Hollow Mimic} (diluted expansion, verbose but shallow; e.g., Deepseek-R1-Qwen-7B), and the \textbf{Lazy Guesser} (saturation/collapse, minimal effort; e.g., Qwen2.5-32B-Instruct). This taxonomy reveals two successful pathways—adaptive strategies that optimize either efficiency or exploration—and two failure modes characterized by unproductive verbosity or cognitive saturation or collapse.

Further, we analyze the impact of inference mode, training strategy, and model scale on these behaviors. Our results reveal that while Long CoT~\citep{openai2024learning, deepseekai2025deepseekr1incentivizingreasoningcapability} generally yields higher Logical Depth ($W$) than Short CoT~\citep{wei2022chain} under similar settings, Short CoT proves capable of reaching substantial depths, often rivaling Long CoT. Specifically, we find that: (1) when trained with reasoning tasks, Short CoT can approach the depth of Long CoT, particularly when enhanced with reflection mechanisms; (2) Distilling Long CoT capabilities into small models often leads to ``behavioral mimicry'', extending $t$ without increasing $W$ due to intrinsic capacity constraints; and (3) mixing long and short CoT data risks disrupting model strategies, often causing premature saturation and collapse. 



Our contributions are threefold: 
1. We propose the first neuro-symbolic evaluation framework in FOL for deterministic, non-intrusive reasoning quantification.
2. By quantifying efficiency, we identify four \textbf{behavioral prototypes} and diagnose critical failure modes.
3. We challenge the assumption that deep reasoning requires extensive token consumption~\cite{chen2025reasoning}. 

We release ReEfBench (including data and evaluation methods) at anonymous.4open.science/r/LoG-1AD8/.

\section{Related Work}

\begin{table*}[h]
    \centering
    \small
    \begin{tabular}{@{} l|ccc | cccc @{}} 
        \toprule
        \multirow{2}{*}{\textbf{Framework}} 
        & \multicolumn{3}{c}{\textbf{Dataset Property}} 
        & \multicolumn{4}{c}{\textbf{Evaluation Property}} \\
        \cmidrule(lr){2-4} \cmidrule(lr){5-8}
        & \textbf{Scalable} & \textbf{FOL} & \textbf{Logic-Only} 
        & \textbf{LogDepth} & \textbf{BehProc} & \textbf{Interp} & \textbf{NonIntr} \\ 
        \midrule
        \textbf{ReEfBench (Ours)} & \checkmark & \checkmark & \checkmark & \checkmark & \checkmark & \checkmark & \checkmark \\ 
        FOLIO~\citep{han-etal-2024-folio} & \xmark & \checkmark & \xmark & -- & -- & -- & -- \\ 
        ProntoQA~\citep{saparov2023language}      & $\sim$ & \checkmark & \checkmark & \xmark & \xmark & \checkmark & \xmark \\ 
        ZebraLogic~\citep{lin2025zebralogic}    & \xmark & \xmark & \checkmark & \checkmark & \xmark & \checkmark & \xmark \\ 
        LogiNumSynth~\citep{liu2025loginumsynth}  & \checkmark & \checkmark & \checkmark & \checkmark & \xmark & \checkmark & \xmark \\
        Sys2Bench~\citep{parashar2025inference}     & -- & -- & -- & \xmark & \checkmark & \checkmark & \checkmark \\ 
        CognitiveBehaviors~\citep{gandhi2025cognitive} & -- & -- & -- & \xmark & \checkmark & \checkmark & \xmark \\ 
        Roscoe~\citep{golovneva2023roscoe}        & -- & -- & -- & \checkmark & $\sim$ & $\sim$ & \checkmark \\ 
        ReCEval~\citep{prasad-etal-2023-receval}       & -- & -- & -- & \checkmark & \xmark & \xmark & \checkmark \\ 
        \bottomrule
    \end{tabular}
    \caption{Comparison of existing frameworks/datasets against our method across dataset properties (Scalable, FOL: First-Order Logic, Logic-Only) and evaluation properties (LogDepth: Logical Depth, BehProc: Behavioral Process, Interp: Interpretable, NonIntr: Non-intrusive). The symbol $\sim$ indicates partial attainment.}
    \label{tab:data_comparison}
\end{table*}

\paragraph{Long CoT and Evaluation Challenges.} Recent reasoning LLMs leverage test-time scaling to tackle complex problems by allocating increased computation to reasoning steps~\citep{openai2024learning}. While Long CoT effectively improves reasoning capabilities~\citep{openai2024learning,deepseekai2025deepseekr1incentivizingreasoningcapability}, extended chains often manifest as ``overthinking''—redundant verification or verbosity with minimal accuracy gains~\citep{chen2025do, peng2025revisitingoverthinkinglongchainofthought}. This motivates a comprehensive evaluation of the CoT process to provide the insights necessary for diagnosing these inefficiencies.

\begin{figure}[h]
    \centering
    \includegraphics[width=0.98\columnwidth]{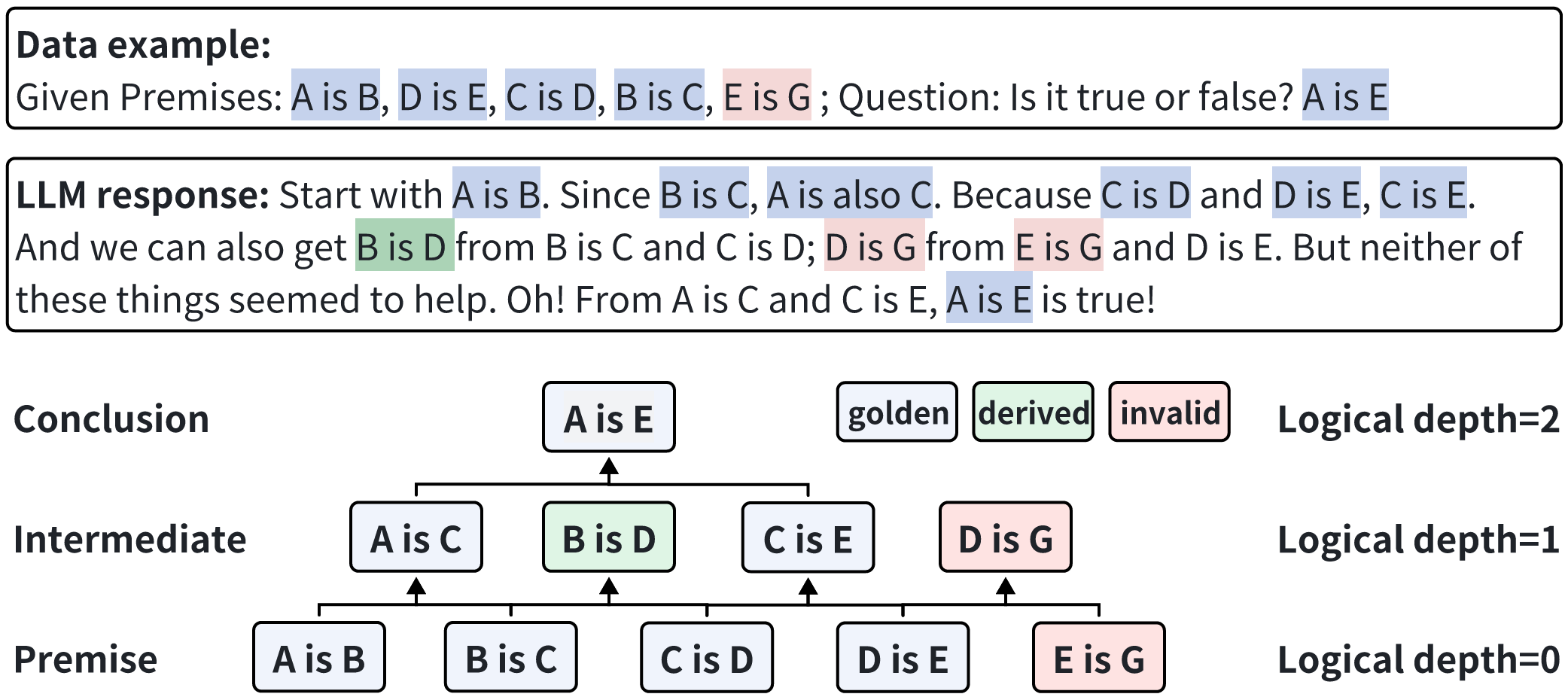}
    \caption{Example of our dataset: Premise, Intermediate and Conclusion. Dataset Complexity = max(Logical depth) = 2.} 
    \label{fig:data_example}
\end{figure}

Current evaluation methods are insufficient for diagnosing these process-level discrepancies. Human annotation is prohibitively expensive, while ``LLM-as-a-Judge''~\citep{fu2024gptscore} approaches suffer from biases and noisy judgments~\citep{lee2025correctly}. Similarly, early automated metrics largely rely on pure rule-based statistics or uninterpretable scores~\citep{golovneva2023roscoe,prasad-etal-2023-receval}, failing to capture the hallmarks of System 2 reasoning like exploration and reflection~\citep{chen2025reasoning}. In logical reasoning, metrics often lack granularity or flexibility. As the most prevailing process metric, ``Validity''~\citep{saparov2023language,zhou2025dissecting,wei-etal-2025-satbench} serves merely as a binary strict accuracy metric. While some step-level evaluations exist, they are typically intrusive~\citep{saparov2023language,lin2025zebralogic,liu2025loginumsynth}, enforcing rigid output formats that limit model flexibility. 
Furthermore, recent studies on exploration and reflection lack a holistic framework to assess logical depth and cognitive behaviors simultaneously~\citep{parashar2025inference,nie2024evolve,heyman2025evaluating,xie2025logicrlunleashingllmreasoning}.
In contrast, by combining the flexibility of neural parsing with the rigor of symbolic verification, our approach mitigates the brittleness of rule-based metrics and the stochasticity of LLM judges, enabling non-intrusive, comprehensive assessment of both logical depth and cognitive behaviors. The systematical comparison is in Table~\ref{tab:data_comparison}.


\paragraph{Neural Symbolic Approaches.} Neuro-symbolic AI converges connectionist pattern recognition with symbolic rigor. In the LLM era, this paradigm typically enhances reasoning capabilities by parsing natural language into formal logic (e.g., SQL, FOL) for rigorous execution by external solvers~\citep{olausson2023linc,pan-etal-2023-logic,wang2025adaptiveselectionsymboliclanguages}. 
In this work, we repurpose this architecture for evaluation. Instead of aiding the model in solving tasks, we employ small LLMs solely to parse semantic variability into structured forms, delegating metric calculation to a symbolic module. 

\section{Method}
\label{sec:methodology}
As illustrated in Figure~\ref{fig:head_fig}, ReEfBench consists of five main stages: scalable dataset construction (Phase A; Section~\ref{subsec:data_construction}), target LLM generation (Phase B; Section~\ref{subsec:parser}), response decomposition/parsing (Phase C; Section~\ref{subsec:parser}), structured processing of the decomposed reasoning nodes (Phase D; Section~\ref{subsec:rules}) and metrics of reasoning process (Phase E; Section~\ref{subsec:metrics}). 

\subsection{Reasoning Data Construction}
\label{subsec:data_construction}


\paragraph{Dataset Formulation.} As illustrated in Figure~\ref{fig:data_example}, each instance in our dataset consists of a set of \textit{premises}, a target \textit{conclusion}, and a step-by-step \textit{golden solution}. The target model is presented with the premises and required to verify the truthfulness of the conclusion. Structurally, the basic \textbf{logical nodes} follow the form ``$x$ is $A$'', representing ``all instances of $x$ are members of $A$''. To support complex reasoning, we employ \emph{Modus Ponens} (e.g., inferring ``Rex is a vertebrate'' from ``Rex is a dog'' and ``All dogs are vertebrates'') as the atomic inference rule, extended with conjunction ($\land$) and disjunction ($\lor$), yielding the four deduction rules summarized in Table~\ref{tab:deduction_rules} (Appendix~\ref{app:deduction_rules}).


\paragraph{Scalable Data Construction.} As outlined in Phase A of Figure~\ref{fig:head_fig}, our data construction process is parameterized by the dataset \textit{complexity level} $C$ and sample size $N$. Referring to Figure~\ref{fig:data_example}, we define the \textit{complexity level} $C$ as the \textit{Logical Depth} of the conclusion node, corresponding to the maximum number of inference layers in the golden solution. 
To construct these samples, we employ a stochastic backward-chaining process: we initialize a random conclusion node and recursively expand the graph backward by sampling valid deduction rules to generate the necessary valid premises until the reasoning chain reaches the target logical depth $C$ (detailed in Algorithm~\ref{alg:logic_graph_simple}, Appendix~\ref{app:Algorithm_detail}).
Upon completion, the leaf nodes serve as the \textit{premises}, the root node acts as the \textit{conclusion}, and the remaining nodes constitute the \textit{intermediate steps} of the golden solution in Figure~\ref{fig:data_example}. Finally, to increase task difficulty, we augment each instance with $C$ extra invalid premises (distractors).


\subsection{LLM response and LLM-based extraction}
\label{subsec:parser}
The reasoning data generated in Section~\ref{subsec:data_construction} (Phase A in Figure~\ref{fig:head_fig}) is passed to target LLMs to obtain responses (Phase B). The details of the prompt and hyper-parameters are shown in Appendix~\ref{app:exp_setting_detail}. The responses will then be processed through the pipeline below (Phase C).


\paragraph{Decomposition and Parsing.}
Given a response, we first decompose it into individual sentences based on punctuation marks (such as periods, question marks, etc.) to significantly reduce the difficulty of processing for LLMs. Subsequently, we use an LLM-based parser to identify the \textbf{type of each sentence}, classifying them as either \emph{normal} or \emph{reflection}~\citep{xie2025logicrlunleashingllmreasoning}, and to extract the logical nodes within the sentences, determining the \textbf{type of logical nodes} as either \emph{actual} or \emph{planning}, representing logical nodes that denote actual events and planning-related logical nodes, respectively. As for the example in Figure~\ref{fig:head_fig}, ``The statement x is bablpus or babypus is true'' is classified as an \textit{actual} node, denoted ``x is bablpus or babypus''. Similarly, representative examples of \emph{planning} and \emph{reflection} are also explicitly illustrated in Phase C of Figure~\ref{fig:head_fig}. The prompts and code for the LLM parser can be found in our open-source repository.

\paragraph{Parser Validation.}
To ensure high reliability, we construct a validation set of 1,000 samples from 5 models manually labeled by three independent human annotators. The annotators demonstrate exceptional consistency with a 95.1\% agreement rate. On this high-quality dataset, the parser based on Qwen2.5-32B-Instruct~\citep{qwen2025qwen25technicalreport} achieves an F1 of 94.3\%, confirming its effectiveness in replicating human classification behavior.

\subsection{Rule-based Node Processing}
\label{subsec:rules}
As in Figure~\ref{fig:head_fig} Phase C, we apply a rule-based method to classify the correctness of each \emph{actual} node produced in Section~\ref{subsec:parser} (using Algorithm~\ref{alg:is_provable} in Appendix~\ref{app:Algorithm_detail}). 
Concurrently, a separate procedure calculates the \emph{logical depth} for arbitrary actual nodes with backward chaining. For nodes explicitly present in the golden solution, we directly retrieve their depth (Figure~\ref{fig:data_example}). For correct nodes absent from the golden solution, we calculate their depth by identifying the deepest nodes required to derive them (detailed in Algorithm~\ref{alg:get-equivalent-depth}, Appendix~\ref{app:Algorithm_detail}). For instance, following Figure~\ref{fig:data_example}, although ``B is D'' is not in the golden solution, it is assigned a \textit{logical depth} of 1 based on its logical antecedents.


For \emph{planning} nodes and \emph{reflection} sentences, we assess their functional utility by examining whether subsequent nodes (within a fixed contextual window) exhibit measurable changes in logical depth or breadth. For instance, we validate a planning node if it is followed by a corresponding actual inference, whereas a reflection sentence is considered effective only if it yields a novel or deeper logical node within a local window (e.g., within the next 5 sentences).


\subsection{Evaluation Metrics}
\label{subsec:metrics}
We aggregate the node classifications and attributes derived in Section~\ref{subsec:rules}, alongside basic statistics (e.g., token counts), into six interpretable diagnostic scores (visualized via the radar chart in Figure~\ref{fig:head_fig}). Specifically:
(1) \textit{Logical Depth} ($S_\textit{ld}$) reflects reasoning capability via the achieved valid logical depth;
(2) \textit{Cost} ($S_\textit{cost}$) captures computational consumption, combining total token count and frequency of reflection/planning steps;
(3) \textit{Exploration} ($S_\textit{exp}$) counts unique, correct logical nodes explored;
(4) \textit{Efficiency} ($S_{\textit{eff}}$) integrates token efficiency (tokens per depth increment) and effective span (normalized position where new node generation stops);
(5) \textit{Coherence} ($S_\textit{coh}$) assesses whether meta-cognitive steps (planning/reflection) lead to actual logical progress;
(6) \textit{Redundancy} ($S_\textit{red}$) quantifies repetition at sentence and node levels.
Given that these dimensions often involve disparate units (e.g., $S_{\mathit{cost}}$ combines raw token counts and step frequencies), we require a unified scale for aggregation. Therefore, we perform max-normalization to map all raw sub-metrics into the $[0,1]$ range. The final score for each dimension is computed as the average of its normalized components. Complementarily, we also utilize the interpretable raw values for fine-grained analysis. For further details on calculating these six metrics and the raw metrics, please refer to Appendix~\ref{app:full_metrics}.

\begin{table*}[t]
    \centering
    \small
    \begin{tabular}{c|l|c|c|cc|cccc|cc}
    \toprule
    \multirow{2}{*}{\#} & \multirow{2}{*}{Model} & \multicolumn{2}{c|}{\textbf{Classification}} & \multicolumn{2}{c|}{\textbf{Core Metrics}} & \multicolumn{4}{c|}{\textbf{Diagnostic Metrics}} & \multicolumn{2}{c}{\textbf{Raw Stats.}} \\
    \cmidrule(lr){3-4} \cmidrule(lr){5-6} \cmidrule(lr){7-10} \cmidrule(lr){11-12}
    & & Category & $S_\textit{c}$ & $S_\textit{ld}$ & $S_\textit{cost}$ & $S_\textit{exp}$ & $S_\textit{eff}$ & $S_\textit{coh}$ & $S_\textit{red}$ & \textbf{Depth} & \textbf{Token(k)} \\
    \midrule
    1 & Qwen3-235B-thinking & DeepWanderer & 1.00 & 1.00 & 0.88 & 1.00 & 0.47 & 0.41 & 0.77 & 10.54 & 16.8 \\
    \midrule
    2 & Qwen3-235B-Instruct & \multirow{5}{*}{EffectiveSolver} & 0.82 & 0.95 & 0.37 & 0.83 & 0.59 & 0.28 & 0.62 & 9.96 & 3.4 \\
    3 & DeepSeek-R1 & & 0.80 & 0.86 & 0.41 & 0.34 & 0.59 & 0.58 & 0.48 & 9.04 & 3.7 \\
    4 & Claude-Opus-4.5.long & & 0.80 & 0.97 & 0.37 & 0.47 & 0.60 & 0.42 & 0.28 & 10.27 & 3.5 \\
    5 & Qwen3-235B.long & & 0.67 & 0.81 & 0.46 & 0.29 & 0.57 & 0.31 & 0.55 & 8.54 & 4.1 \\
    6 & Claude-Opus-4.5.short & & 0.33 & 0.74 & 0.24 & 0.62 & 0.70 & 0.47 & 0.37 & 7.82 & 1.4 \\
    \midrule
    7 & DS-R1-Qwen-7B & \multirow{4}{*}{HollowMimic} & 1.00 & 0.46 & 0.49 & 0.01 & 0.47 & 0.35 & 0.62 & 4.80 & 6.0 \\
    8 & Qwen3-14B.long & & 0.39 & 0.47 & 0.39 & 0.09 & 0.54 & 0.34 & 0.57 & 4.90 & 3.4 \\
    9 & QwQ-32B & & 0.34 & 0.68 & 0.61 & 0.14 & 0.48 & 0.32 & 0.62 & 7.12 & 5.7 \\
    10 & Qwen3-32B.long & & 0.29 & 0.58 & 0.35 & 0.24 & 0.56 & 0.33 & 0.52 & 6.14 & 2.7 \\
    \midrule
    11 & Qwen2.5-32B-Inst & \multirow{4}{*}{LazyGuesser} & 1.00 & 0.28 & 0.16 & 0.03 & 0.55 & 0.07 & 0.59 & 2.90 & 0.7 \\
    12 & Qwen3-4B.short & & 1.00 & 0.45 & 0.17 & 0.02 & 0.67 & 0.11 & 0.54 & 4.70 & 0.7 \\
    13 & Qwen3-235B.short & & 0.74 & 0.54 & 0.20 & 0.12 & 0.70 & 0.43 & 0.53 & 5.70 & 1.0 \\
    14 & Qwen3-4B.long & & 0.62 & 0.42 & 0.29 & 0.03 & 0.55 & 0.39 & 0.48 & 4.38 & 1.7 \\
    \midrule
    \multicolumn{2}{c|}{\multirow{4}{*}{\textit{Category Avg (weighted)}}} & \multicolumn{2}{c|}{DeepWanderer} & \textbf{1.00} & \textbf{0.88} & \textbf{1.00} & 0.47 & 0.41 & \textbf{0.77} & \textbf{10.54} & \textbf{16.8} \\
    \multicolumn{2}{c|}{} & \multicolumn{2}{c|}{EffectiveSolver} & 0.86 & 0.37 & 0.42 & 0.60 & 0.40 & 0.46 & 9.40 & 3.1 \\
    \multicolumn{2}{c|}{} & \multicolumn{2}{c|}{HollowMimic} & 0.52 & 0.45 & 0.09 & 0.50 & \textbf{0.42} & 0.60 & 5.70 & 4.2 \\
    \multicolumn{2}{c|}{} & \multicolumn{2}{c|}{LazyGuesser} & 0.45 & 0.21 & 0.09 & \textbf{0.62} & 0.25 & 0.51 & 5.03 & 1.2 \\
    \bottomrule
    \end{tabular}
    \caption{Model classification results based on K-means clustering, showing category assignments, confidence scores ($S_\textit{c}$), core metrics ($S_\textit{ld}$, $S_\textit{cost}$), and diagnostic metrics ($S_\textit{exp}$, $S_\textit{eff}$, $S_\textit{coh}$, $S_\textit{red}$) for representative models. The rightmost "Raw Stats." columns are included for reference, where Depth represents the Average Logical Depth (max 11 in this dataset) and Token.(k) denotes the Token Count (in thousands). Category averages are weighted by confidence scores ($S_\textit{c}$) and calculated based on the full set of 25 models (Table~\ref{tab:full_model_classification}).}
    \label{tab:model_classification}
\end{table*}

\section{Experiments}
\label{sec:experiment}
We conduct extensive experiments over ReEfBench, with representative commercial and open-source models to evaluate their reasoning processes. We show main results in Section~\ref{subsec:main_res}, identify four distinct behavioral prototypes in Section~\ref{subsec:prototypes}, and subsequently investigate the dynamic trajectories of model behaviors across varying complexity levels in Section~\ref{subsec:dynamic_trajectory}.

\subsection{Experimental Settings}
\label{subsec:exp-setting}
We utilize the scalable dataset construction method from Section~\ref{subsec:data_construction} to generate evaluation sets spanning Complexity Levels 3 to 11, with 100 samples per level.  
Our evaluation suite encompasses 25 diverse LLMs, including proprietary SOTA models (e.g., Claude-4.5 series~\citep{anthropic2025opus, anthropic2025sonnet}) and open-weights models (e.g., Qwen series~\citep{qwen2025qwen25technicalreport, yang2025qwen3technicalreport, qwq32b} and DeepSeek-R1 series~\citep{deepseekai2025deepseekr1incentivizingreasoningcapability}), covering both standard instruction-tuned and reasoning-enhanced paradigms. For nomenclature, we append .long or .short suffixes to models with distinct thinking or non-thinking modes (e.g., Qwen3-32B.long denotes the thinking mode), whereas single-mode models retain their original abbreviations. Detailed model settings can be found in Table~\ref{tab:model_training_methods} in Appendix~\ref{app:model_details}.

\subsection{Main Results}
\label{subsec:main_res}
To capture model behaviors under maximum stress, we focus exclusively on the most challenging subset (Complexity Level 11). In Table~\ref{tab:model_classification}, we report the statistics of the six metrics defined in Section~\ref{subsec:metrics} across 14 representative models, where we observe substantial behavioral divergence. Frontier reasoning models nearly reach the maximal logical depth of 11, yet their behaviors diverge markedly. Qwen3-235B-thinking consumes 16.6k tokens, reflecting exaustive Exploration (1.0), high Redundancy (0.77), and low Efficiency (0.47), whereas Claude-Opus-4.5.long uses much fewer tokens (3.5k) with high Efficiency (0.60) but lower Redundancy (0.28) and exploration (0.47). In smaller-scale ($\leq$32B), models with short CoT (e.g., Qwen2.5-32B-Instruct) generally incur low cost (<2k tokens) but rarely exceed depth 5, while long-CoT models (e.g., QwQ-32B) spend more tokens (3k--6k) yet struggle to match the depth of frontier models, peaking around depth 7.1. Specifications for all 25 evaluated models are provided in Table~\ref{tab:full_model_classification} in Appendix~\ref{app:experiment_details}.

\subsection{The Reasoning Landscape}
\label{subsec:prototypes}
\begin{figure}[h]
    \centering
    \includegraphics[width=0.85\columnwidth]{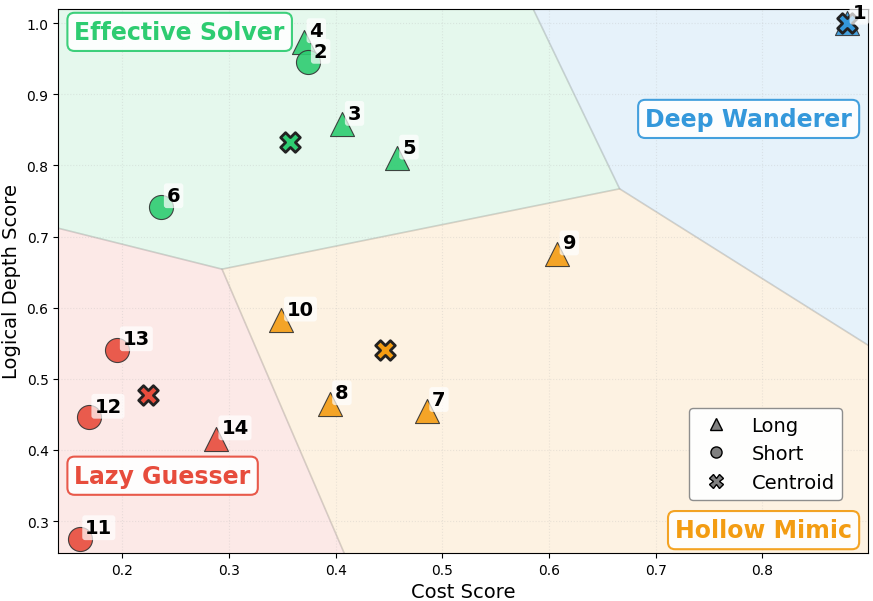}
    \caption{Models plotted by Logical Depth ($S_\textit{ld}$) vs. Cost ($S_\textit{cost}$); triangles = Long CoT, circles = Short CoT, stars = centroids. Model IDs listed in Table~\ref{tab:model_classification}.}   
    \label{fig:kmeans_classification_no_legend}
\end{figure}

To systematize the patterns in Table~\ref{tab:model_classification}, we map the models into a normalized reasoning space defined by Logical Depth ($S_{\mathit{ld}}$, y-axis) and Cost ($S_{\mathit{cost}}$, x-axis). We apply K-means clustering with $k=4$ specifically to capture the \textbf{behavioral prototypes} inherent to the plane's four quadrants, yielding four distinct reasoning archetypes visualized in Figure~\ref{fig:kmeans_classification_no_legend}. We then employ the remaining four diagnostic metrics (Efficiency, Exploration, Coherence, Redundancy) to interpret their internal mechanisms. To accurately characterize these prototypes, we compute the average score of each metric weighted by the confidence score $S_c$ (Table~\ref{tab:model_classification}).

\paragraph{Deep Wanderer}
This cluster forms a distinct behavioral regime characterized by simultaneous high cost ($S_\textit{cost} \approx 0.88$) and deep reasoning ($S_\textit{ld} \approx 1.0$). In our current evaluation, it is uniquely represented by Qwen3-235B-thinking.
While models like QwQ-32B and Qwen3-235B.long approach this cluster in Figure~\ref{fig:kmeans_classification_no_legend}, they lack the extreme depth or cost. Diagnostically, this archetype trades efficiency for coverage: it combines low Efficiency ($0.47$) with high Redundancy ($0.77$) and Exploration ($1.00$). This suggests the model reaches deep reasoning states by rigorously expanding the search space and tolerating redundant steps.

\paragraph{Effective Solver.}
Models in this category, such as Claude-Opus-4.5.long and Qwen3-235B-Instruct, maintain high task performance ($S_\textit{ld}\approx0.86$) but with significantly reduced cost ($S_\textit{cost}\approx0.37$). Unlike Deep Wanderers, they favor direct derivation, exhibiting the lowest Redundancy ($0.46$) and limited search breadth ($0.42$). Surprisingly, this category includes many ``thinking'' models (notably Claude-Opus-4.5.long), challenging the assumption that reasoning enhancement necessitates verbose exploration~\citep{chen2025reasoning}. 

\paragraph{Hollow Mimic.}
Prevalent in smaller reasoning models (such as QwQ-32B), this category is defined by a severe mismatch between effort and outcome: they invest significant cost ($0.45$, 2nd highest) for mediocre logical depth ($0.52$). While their low Efficiency ($0.50$) and high Redundancy ($0.60$) mirror the Deep Wanderer, their Exploration is critically low ($\approx 0.09$), meaning they generate text without expanding the logical search radius. Crucially, these models maintain high Process Coherence ($0.42$) comparable to successful solvers. This reveals a ``performative reasoning'' failure: explicit planning and reflection behaviors are correctly triggered and structured, but fail to translate into genuine logical progress.

\paragraph{Lazy Guesser.}
This cluster includes standard instruction models (Qwen2.5-32B-Instruct) and small reasoning models (Qwen3-4B.long) that fail at high complexities. They exhibit the lowest cost ($0.21$). Although their Efficiency score appears high ($0.61$), this is an artifact of brevity rather than competence. Counter-intuitively, their Redundancy is notable ($0.51$), indicating that ``lazy'' failure is rarely a concise refusal; instead, these models often get stuck in restatements, unable to propel the reasoning forward.

\subsection{Dynamic Trajectories: Scaling with Complexity}
\label{subsec:dynamic_trajectory}
\begin{figure}[ht]
    \centering
    \includegraphics[width=0.49\textwidth]{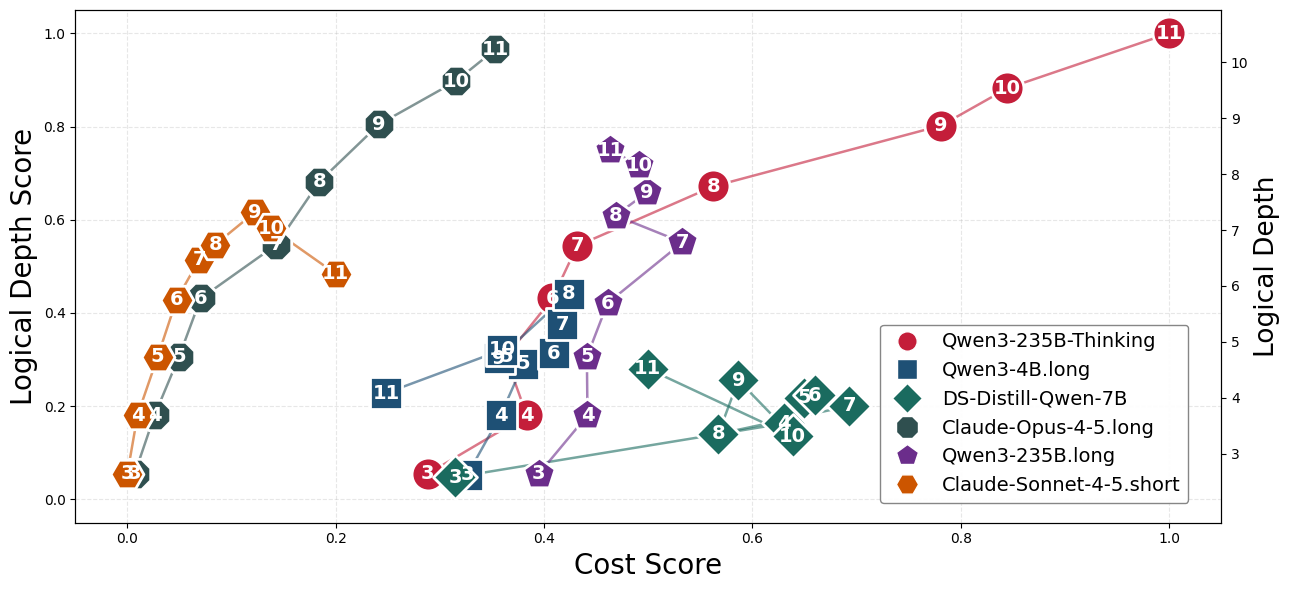}
    \caption{Performance trajectories across varying complexity levels (3--11). The visualization illustrates three distinct behavioral patterns defined in Section 4.3: (1) Adaptive Scaling (e.g., Qwen3-235B-thinking, Claude-Opus-4.5.long); (2) Diluted Expansion (e.g., Claude-Sonnet-4.5.short); and (3) Saturation (e.g., DS-Distill-Qwen-7B) \& Collapse (e.g., Qwen3-4B.long). Complete trajectories for all evaluated models are presented in Figure~\ref{fig:trajectory} in Appendix~\ref{app:experiment_details}.}
    \label{fig:performance_with_scale}
\end{figure}

We analyze the performance trajectories of models across varying complexity levels ($C=3$ to $11$). Figure~\ref{fig:performance_with_scale} illustrates three representative archetypes identified from these trajectories: \textit{Adaptive Scaling} (successful reasoning), and two failure modes—\textit{Saturation/Collapse}~\citep{shojaee2025illusion} (insufficient effort) and \textit{Diluted Expansion} (unproductive verbosity), as introduced below.



\begin{table}[htbp]
    \centering
    \small
    \begin{tabular}{@{}l| c c c c@{}}
        \toprule
        \textbf{Model} & \textbf{Mode} & \textbf{Method} & \textbf{Data} & \textbf{Depth $\Delta$} \\
        \midrule
        QwQ-32B & Long & SFT+RL & R+G & +97.1\% \\
        Qwen2.5-32B-In & Short & SFT & G & - \\
        \midrule
        Claude-Opus-4.5 & Long & - & - & +8.8\% \\
         & Short & - & - & - \\
        \midrule
        Qwen3-235B-Th & Long & - & R+G & +1.6\% \\
        Qwen3-235B-In & Short & - & R+G & - \\
        \midrule
        Qwen3-32B & Long & SFT+RL & R+G & +6.1\% \\
         & Short & SFT+RL & R+G & - \\
        \midrule
        Qwen3-14B & Long & SFT & R+G & +6.5\% \\
         & Short & SFT & R+G & - \\
        \bottomrule
    \end{tabular}
    \caption{Training Paradigm Analysis. We compare Long vs. Short modes at Complexity Level 11. \textbf{Th/In}: Thinking/Instruct. \textbf{Data}: R (Reasoning), G (General). $\Delta$ shows relative gain. Note that QwQ-32B and Qwen2.5-32B-In. represent early model iterations, while the others are current. For raw statistics and additional models, please refer to Table~\ref{tab:model_depth_token_analysis} in Appendix~\ref{app:experiment_details}.}
    \label{tab:model_depth_token_analysis_simple}
\end{table}

\paragraph{Adaptive Scaling.} In an ideal scenario, both cost and logical depth increase as complexity grows. Models exhibiting this pattern correctly identify increased difficulty and generate proportional tokens to resolve it. For instance, Claude-Opus-4.5.long demonstrates high-efficiency solving, while Qwen3-235B-thinking achieves similar success through exhaustive exploration.

\paragraph{Saturation and Collapse.} As complexity rises, some models (e.g., DS-Distill-Qwen-7B) fail to further scale their token generation, causing logical depth to plateau (\textit{Saturation}). More critically, other models (e.g., Qwen3-4B.long) experience \textit{Collapse}~\citep{shojaee2025illusion}, where both token count and logical depth significantly decrease compared to lower complexity levels, indicating a breakdown in reasoning maintenance.

\paragraph{Diluted Expansion.} Other models become verbose without being smarter (e.g., Claude-Sonnet-4.5.short). This pattern involves increased token count in response to difficulty without corresponding gains in logical depth. The implication is clear: for models lacking intrinsic reasoning abilities, scaling output length is necessary but insufficient.

\section{Analysis}
\label{sec:analysis}
Following the identification of behavioral prototypes in Section~\ref{sec:experiment}, we examine the impact of inference mode (Long vs.\ Short CoT), training strategy, and model scale. We apply the $W=Pt$ intuition proposed in the Introduction to interpret these effects, formulating Logical Depth as \textbf{Work ($W$)} and Token Count as \textbf{Time ($t$)} to derive \textbf{Efficiency ($P = W/t$)}.



\subsection{Interpreting Behavioral Trajectories}
\label{sec:physics_interpretation}
Under this lens, the behavioral prototypes map to distinct dynamic trajectories governing the trade-off between Efficiency ($P$) and Cost ($t$). \textbf{Adaptive Scaling} represents the successful equilibrium where models effectively scale effort ($t$) to match rising difficulty (required $W$). Crucially, while successful strategies increase $t$ in response to difficulty, they do so with distinct gradients. The \textit{High-$t$ Strategy} (``Deep Wanderer'') rapidly expands the tokens to accumulate the necessary $W$, often relying on exhaustive search. Conversely, the \textit{High-$P$ Strategy} (``Effective Solver'') also scales $t$, but at a significantly slower rate. 

\begin{figure*}[ht]
    \centering
    \includegraphics[width=0.95\textwidth]{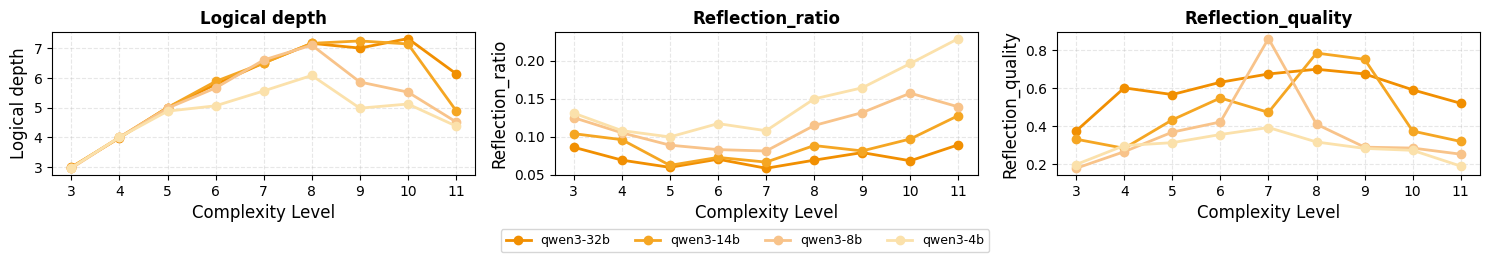}
    \caption{Qwen3-32B vs 14B vs 8B vs 4B. Three subplots compare key reasoning behaviors (logical depth, reflection ratio and quality) across model sizes as problem complexity increases. Smaller distilled models exhibit more sophisticated behaviors, but fail to emulate behavioral efficiency and cannot translate these sophisticated behaviors into deeper reasoning.}
    \label{fig:distill_comparison_simple}
\end{figure*}

In contrast, failure modes represent breakdowns in this scaling mechanism. \textbf{Diluted Expansion} (``Hollow Mimic'') mimics the rapid $t$-expansion of the High-$t$ strategy ($\Delta t \gg 0$) but fails to reach proportional Logical Depth. Alternatively, \textbf{Saturation \& Collapse} (``Lazy Guesser'') occurs when the model fails to scale $t$. Facing increased complexity, the model either retreats to stagnant $t$ or decreasing $t$, causing $W$ to plateau or drop despite the need for greater depth.

\subsection{Long CoT vs. Short CoT}
\label{sec:long_short_cot}
According to Table~\ref{tab:model_depth_token_analysis_simple}, while early High-$t$ models (Long CoT~\citep{chen2025reasoning}) hold a significant advantage in logical depth—e.g., QwQ-32B exceeds Qwen2.5-32B-Instruct by over 97\%—current Short CoT~\citep{wei2022chain}  models are rapidly bridging this divide. The logical depth gap have narrowed to under 10\% in modern iterations, with Qwen3-235B-Instruct trailing its thinking counterpart by a negligible 1.61\%. Table~\ref{tab:model_depth_token_analysis_simple} reveals that this convergence is methodology-agnostic yet correlated with the inclusion of reasoning data. This aligns with \citet{yu2025longshortchainofthoughtmixturesupervised}, who observes that shortening long chains while preserving structure retains reasoning capability. Further, Table~\ref{tab:model_reflection} indicates that high-performing short models consistently incorporate \textit{reflection}, underscoring the critical role of ``Short CoT + Reflection''. Consequently, as short CoT achieve comparable depth at significantly reduced costs, deployment strategies are shifting from functional partitioning (Long for reasoning, Short for general) to efficiency trade-offs.

\subsection{Mixed Training}
\label{sec:mixed_training}
We analyze Qwen3 models under pure ``thinking'' (Long CoT) vs. mixed (Long + Short) training. As visualized in Figure~\ref{fig:performance_with_scale}, while the pure Qwen3-235B-thinking maintains robust \textbf{Adaptive Scaling} by increasing $t$ with complexity, the mixed Qwen3-235B.long suffers premature \textbf{Saturation}, failing to scale token generation significantly earlier. This structural degradation persists across other scales (pure QwQ-32B vs. mixed Qwen3-32B.long, pure Qwen3-30B-thinking vs. mixed Qwen3-30B.long), as detailed in Figure~\ref{fig:seperate_vs_mixed} in Appendix~\ref{app:experiment_details}.

We suggest that incorporating short-CoT data (High-$P$) has the risk of negatively interfering with the High-$t$ mechanism, creating a reluctance to scale computation when necessary. This interference explains why Qwen3-2507~\citep{Qwen3Thinking2507} have abandoned mixed training strategies in favor of specialized tuning.

\subsection{Distillation}
\label{sec:distillation_limits}


We examine the limits of distilling Long CoT (High-$t$) strategies from larger teachers into smaller students using the Qwen3 lineage (14B, 8B, 4B) supervised by a 32B teacher.As visualized in Figure~\ref{fig:distill_comparison_simple}, smaller models (4B, 8B) generate more reflection steps than the teacher at lower difficulties. 

\begin{table}[ht]
    \centering
    \small
    \begin{tabular}{@{}lcc@{}}
        \toprule
        \textbf{Model} & \textbf{Logical Depth} & \textbf{Reflection\%} \\ \midrule
        Claude-Opus-4.5.long & 10.3 & 0.7\% \\ 
        Qwen3-235B-Instruct & 10.0 & 1.8\% \\ 
        Claude-Opus-4-5.short & 7.8 & 0.8\% \\ 
        \midrule
        Qwen3-235b.short & 5.7 & 0.1\% \\ 
        Qwen3-32b.short & 5.7 & 0.0\% \\ 
        Qwen2.5-32B-Instruct & 2.9 & 0.0\% \\ 
        \bottomrule
    \end{tabular}
    \caption{Impact of Reflection on Logical Depth. Comparing Effective Solvers (top) against Lazy Guessers (bottom) reveals that successful models maintain reflection steps even in short reasoning process.}
    \label{tab:model_reflection}
\end{table}

However, this appearance is deceptive. While the \textit{quantity} (frequency) of reflection mimics or exceeds the teacher, the semantic \textit{quality} degrades strictly with model size ($32B > 14B > 8B > 4B$), failing to translate into Logical Depth ($W$). We observe an identical dissociation between frequency and quality in planning behaviors, as detailed in Figure~\ref{fig:distill_comparison} in Appendix~\ref{app:experiment_details}. In contrast, Qwen3-14B marks a clear divergence: it successfully maintains both reflection quality and logical depth aligned with the 32B teacher. This distinct separation indicates that without sufficient parametric capacity (with 14B emerging as the critical threshold), forcing a Long CoT strategy results in ``Diluted Expansion''—the model mimics the form of reasoning to minimize distillation loss without grasping the intrinsic logic.

\section{Conclusion}
\label{sec:conclusion}
We introduce a neuro-symbolic framework that grounds natural language reasoning in FOL to deterministically quantify reasoning efficiency of LLMs. Through this lens, we identify four reasoning prototypes and diagnose their behavioral characteristics. Our analysis reveals that Long CoT is not necessary for deep reasoning. Furthermore, mixing long and short CoT in training risks strategy interference, while distillation often yields behavioral mimicry. Our dataset, ReEfBench, and the evaluation method can be used for efficiency evaluation of the reasoning process.

\section*{Limitations}
We acknowledge three limitations. First, restricting evaluation to First-Order Logic (FOL) prioritizes verification rigor over generality; consequently, our findings may not strictly apply to domains like open-ended QA. Second, our non-intrusive design limits our analysis to verifying the logic of the naturally generated text, rather than forcing the externalization of implicit reasoning steps. Finally, regarding the four reasoning archetypes: while these patterns are real, the boundaries are not absolute. A model's specific category is relative and can be fuzzy; however, by analyzing a diverse set of models, we ensure that the overall classification and statistical trends remain objective.
\bibliography{custom}

\appendix

\section{Methodology Supplement}
\label{app:methodology_supplement}
\subsection{Dataset Details}
\label{app:dataset_detail}
We calculate the accuracy of all the models evaluated with data from Complexity Level 3 to Level 11 (Table~\ref{tab:model_acc_extended}). Despite the fundamental logic relying on simple Modus Ponens combined with and/or operators, the most advanced model (Claude-Opus-4.5.long) exhibits a model accuracy still declines rapidly (0.62) under high complexity. 
For clarity, models bearing the suffixes .long or .short (e.g., Claude-Sonnet-4.5.long, Qwen3-32B.short) denote dual-mode variants of the same model. Models without such suffixes are single-mode versions. Regarding model versions and release dates: the Qwen3-235B-thinking, Qwen3-235B-instruct, Qwen3-30B-thinking, and Qwen3-30B-instruct models are specialized single-mode variants derived from their respective base models (Qwen3-235B and Qwen3-30B) and correspond to the 2507 training snapshot. The two Claude-Opus-4.5 variants were released on November 1 (1101), while the Claude-Sonnet-4.5 variants date from September 29 (0929). DeepSeek-R1 was released on May 28 (0528).

\begin{table*}[h]
\centering
\resizebox{\textwidth}{!}{%
\begin{tabular}{lccccccccc}
\toprule
\textbf{Model} & \textbf{Level\_3} & \textbf{Level\_4} & \textbf{Level\_5} & \textbf{Level\_6} & \textbf{Level\_7} & \textbf{Level\_8} & \textbf{Level\_9} & \textbf{Level\_10} & \textbf{Level\_11} \\
\midrule
Claude-Sonnet-4.5.long & 100\% & 100\% & 100\% & 100\% & 99\% & 99\% & 98\% & 92\% & 62\% \\
Claude-Sonnet-4.5.short & 100\% & 99\% & 99\% & 94\% & 85\% & 56\% & 66\% & 45\% & 33\% \\
Claude-Opus-4.5.long & 100\% & 100\% & 100\% & 100\% & 97\% & 99\% & 98\% & 87\% & 63\% \\
Claude-Opus-4.5.short & 100\% & 100\% & 100\% & 100\% & 95\% & 74\% & 56\% & 49\% & 32\% \\
Deekseek-R1 & 100\% & 100\% & 100\% & 99\% & 97\% & 79\% & 70\% & 25\% & 17\% \\
Qwen3-235B.long & 98\% & 96\% & 97\% & 92\% & 90\% & 73\% & 39\% & 25\% & 14\% \\
Qwen3-235B.short & 97\% & 95\% & 92\% & 62\% & 74\% & 47\% & 23\% & 15\% & 7\% \\
Qwen3-30B.long & 100\% & 91\% & 90\% & 54\% & 28\% & 9\% & 2\% & 3\% & 0\% \\
Qwen3-30B.short & 98\% & 83\% & 69\% & 41\% & 24\% & 9\% & 6\% & 3\% & 5\% \\
Qwen3-30B-thinking & 78\% & 72\% & 92\% & 79\% & 48\% & 33\% & 9\% & 3\% & 0\% \\
Qwen3-30B-instruct & 92\% & 91\% & 86\% & 69\% & 59\% & 43\% & 17\% & 14\% & 7\% \\
Qwen3-235B-thinking & 94\% & 99\% & 100\% & 100\% & 94\% & 91\% & 82\% & 63\% & 51\% \\
Qwen3-235B-instruct & 98\% & 96\% & 98\% & 98\% & 99\% & 95\% & 82\% & 49\% & 30\% \\
Qwen2.5-32B-instruct & 93\% & 54\% & 40\% & 25\% & 15\% & 11\% & 8\% & 6\% & 5\% \\
QwQ-32B & 99\% & 95\% & 91\% & 84\% & 57\% & 26\% & 10\% & 6\% & 1\% \\
DS-R1-Qwen-32B & 99\% & 92\% & 93\% & 74\% & 35\% & 16\% & 10\% & 10\% & 9\% \\
DS-R1-Qwen-7B & 56\% & 29\% & 32\% & 9\% & 7\% & 3\% & 1\% & 1\% & 3\% \\
Qwen3-32B.short & 93\% & 93\% & 94\% & 62\% & 68\% & 40\% & 16\% & 13\% & 11\% \\
Qwen3-32B.long & 98\% & 99\% & 100\% & 79\% & 74\% & 50\% & 20\% & 11\% & 4\% \\
Qwen3-14B.long & 97\% & 94\% & 95\% & 80\% & 69\% & 42\% & 27\% & 4\% & 3\% \\
Qwen3-14B.short & 90\% & 84\% & 75\% & 43\% & 49\% & 32\% & 19\% & 15\% & 10\% \\
Qwen3-8B.long & 95\% & 93\% & 89\% & 76\% & 52\% & 33\% & 8\% & 3\% & 0\% \\
Qwen3-8B.short & 81\% & 72\% & 68\% & 60\% & 35\% & 19\% & 11\% & 6\% & 0 \\
Qwen3-4B.long & 94\% & 92\% & 77\% & 28\% & 22\% & 9\% & 0\% & 2\% & 0\% \\
Qwen3-4B.short & 90\% & 70\% & 54\% & 37\% & 35\% & 16\% & 13\% & 6\% & 2\% \\
\bottomrule
\end{tabular}%
}
\caption{Model accuracy across difficulty levels (Level 3 to Level 11), showing performance degradation as complexity increases, with notable variations among models and configurations (e.g., .long and .short).}
\label{tab:model_acc_extended}
\end{table*}

\subsection{Table for deduction rules}
\label{app:deduction_rules}
Table~\ref{tab:deduction_rules} shows the deduction rules we apply to construct our datasets. 
\begin{table*}[t]
\small
\centering
\begin{tabular}{l c l}
\toprule
\textbf{Deduction rule} & \textbf{Formal definition} & \textbf{Natural language example} \\
\midrule
Modus Ponens
&
$\dfrac{f(a)\quad \forall x \,(f(x)\vdash g(x))}{g(a)}$
&
Alex is a cat. All cats are carnivores.  
Alex is a carnivore. \\[6pt]

Conjunction Introduction
&
$\dfrac{A \quad B}{A \land B}$
&
Alex is a cat. Alex is orange.  
Alex is a cat and orange. \\[6pt]

Conjunction Elimination
&
$\dfrac{A \land B}{A}$
&
Alex is a cat and orange.  
Alex is a cat. \\[6pt]

Disjunction Introduction
&
$\dfrac{A}{A \lor B}$
&
Alex is a cat.  
Alex is a cat or orange. \\
\bottomrule
\end{tabular}
\caption{Four deduction rules used in our dataset, derived from Modus Ponens with logical conjunction ($\land$) and disjunction ($\lor$). Adapted from \citet{saparov2023testing}.}
\label{tab:deduction_rules}
\end{table*}

\subsection{Main Concepts for Methodology}
\label{app:main_concept}
\paragraph{Logical Graph.}
A \emph{Logical Graph} is a directed graph $\mathcal{G} = (\mathcal{V}, \mathcal{E}, \tau)$, where:
\begin{itemize}
    \item $\mathcal{V}$ is a set of \emph{statements}, each has form $X \vdash Y$;
    \item $\mathcal{E}$ encodes inference dependencies (i.e., an edge $(u, v)$ exists if $v$ is derived using $u$);
    \item $\tau$ assigns each node a type label from \textsc{Premise}, \textsc{Derived}, \textsc{Planning}, or \textsc{Hallucination}.
\end{itemize}

A \emph{logical complete tree} is a tree in which all leaf-to-root paths have equal length. The \emph{canonical LoG}, denoted $\mathcal{G}^\star = (\mathcal{V}^\star, \mathcal{E}^\star, \tau^\star)$, is the unique logical complete tree that (i) contains all and only sound inferences from $\mathcal{P}$, and (ii) excludes all \textsc{Hallucination} nodes.

\paragraph{Logical Depth ($D$).}
We define logical depth at the statement level. Let $\mathcal{S}_0 = \mathcal{P}$. The depth of a statement $s$ is the smallest $k$ such that $s$ can be derived in $k$ inference steps:
\[
\text{depth}(s) = 
\begin{cases}
0 & \text{if } s \in \mathcal{P}, \\
k & \text{if } s \in \mathcal{S}_k \text{ and } s \notin \bigcup_{i=0}^{k-1} \mathcal{S}_i, \\
-1 & \text{if } \mathcal{P} \not\vdash s \text{ (i.e., hallucination)}.
\end{cases}
\]
For non-canonical graphs, we compute depth recursively. Define $\mathcal{C}_0 = \mathcal{P}$, and for $k \geq 0$:
\[
\mathcal{C}_{k+1} = \left\{ 
Z \;\middle|\;
\begin{aligned}
& \exists m \geq 1,\ \exists Y_1, \dots, Y_m \text{ such that} \\
& (Y_1 \land \dots \land Y_m) \vdash Z \text{ a valid rule}, \\
& Y_1, \dots, Y_m \in \bigcup_{t=0}^{k} \mathcal{C}_t, \\
& \max_{1 \leq \ell \leq m} \min \{ t \mid Y_\ell \in \mathcal{C}_t \} = k
\end{aligned}
\right\}.
\]
Then $\text{depth}(s) = k$ iff $s \in \mathcal{C}_k \setminus \bigcup_{t<k} \mathcal{C}_t$.
Importantly, depth is \emph{unique} for any sound statement, regardless of the derivation path.

Moreover, the number of premises required to derive a conclusion at depth $d$ scales as $n_\text{premise} \sim m^d$, where $m$ is a constant branching factor determined by the dataset.

\paragraph{Logical Breadth ($B$ and $B^\star$).}
We define \emph{logical breadth} as the size of the reachable logical closure:
\[
B = |\{ s \mid \mathcal{P} \vdash s \}|,
\]
i.e., the total number of distinct statements derivable from $\mathcal{P}$. The \emph{minimal necessary breadth} $B^\star$ is the size of the smallest subset of this closure that suffices to derive the target conclusion, which captures the essential reasoning scope required for a given task.

\paragraph{Reasoning Cost ($t$).}
We proxy computational cost by the number of tokens in the model’s CoT response, denoted $t$. While imperfect, this provides a practical, observable measure of reasoning effort. 

\subsection{Definition of Node Types}
\label{app:def_node}
Table~\ref{tab:node_def} shows the definition of all node types.

\begin{table*}[t]
    \centering
    \small
    \renewcommand{\arraystretch}{1.3}
    \setlength{\tabcolsep}{8pt}     
    \begin{tabular}{l l l p{8cm}}
        \toprule
        \textbf{Category} & \textbf{Sub-type} & \textbf{Formal Definition} & \textbf{Description} \\
        \midrule
        
        \textbf{Planning} & - & - & Meta-cognitive statements outlining strategy (e.g., ``Let us assume...'') without asserting a logical conclusion. \\
        \midrule
        
        \multirow{3}{*}{\textbf{Actual}} 
        
        & \textbf{Premise} & $s \in \mathcal{P}$ & 
        Fundamental facts explicitly provided in the problem description. \\
        
        & \textbf{Derived} & $\mathcal{P} \vdash s$ & 
        Logically valid deductions implied by $\mathcal{P}$\\
        
        & \textbf{Hallucination} & $\mathcal{P} \nvdash s$ & 
        Statements that cannot be logically inferred from $\mathcal{P}$\\
        
        \bottomrule
    \end{tabular}
    \caption{Taxonomy of reasoning node types. Nodes are first classified into \textit{Planning} or \textit{Actual} steps by the parser. Actual steps are further categorized based on their logical validity and relationship to the canonical ground truth graph $\mathcal{G}^\star$ and premise set $\mathcal{P}$.}
    \label{tab:node_def}
\end{table*}

\subsection{Algorithm Details}
\label{app:Algorithm_detail}

\paragraph{\texttt{data\_generation}}

\begin{algorithm}[t]
\caption{Logical Reasoning Graph Generation}
\label{alg:logic_graph_simple}
\begin{algorithmic}[1]
\Require 
Maximum reasoning depth $H$; \\
Element vocabulary $\mathcal{E}$; \\
Deduction rules $\{\text{MP},\text{CE},\text{CI},\text{DI}\}$; \\
Hard mode flag $\textsc{Hard}$
\Ensure 
A logical reasoning graph $G$

\State Initialize unused elements $\mathcal{E}' \leftarrow \mathcal{E}$

\State Sample an initial conclusion $c_0$ using $1\!\sim\!3$ elements from $\mathcal{E}'$
\State Initialize graph $G$ with root node $c_0$

\For{$d = 1$ to $H$}
    \For{each leaf conclusion $c$ at depth $d$}
        \State Select a valid deduction rule $r$
        \State Generate premises $\mathcal{P} \leftarrow r(c)$
        \State Add $(\mathcal{P} \Rightarrow c)$ to $G$
    \EndFor
\EndFor

\If{\textsc{Hard}}
    \State Add irrelevant premises as distractors at maximum depth
\EndIf

\State \Return $G$
\end{algorithmic}
\end{algorithm}

This method automatically generates multi-hop logical reasoning graphs using a fixed set of deductive rules (Modus Ponens, Conjunction Elimination, Conjunction Introduction, Disjunction Introduction). Starting from a randomly constructed conclusion, the algorithm recursively expands the graph backward by selecting admissible inference rules under structural constraints to avoid trivial or cyclic reasoning. Each graph is generated via breadth-first expansion up to a predefined depth, and an optional hard mode augments the premises with logically irrelevant distractors. The resulting graphs are finally converted into question–answer pairs for evaluating logical reasoning capabilities.

\paragraph{\texttt{is\_provable}}
\begin{algorithm}[h]
  \small
  \caption{Backward Chaining Proof Search}
  \label{alg:is_provable}
  \begin{algorithmic}[1]
  \Function{IsProvable}{$\tau, \Pi, V, d, t_0$}
      \State \textbf{Input:} target $\tau$, premises $\Pi$, visited $V$, depth $d$, start time $t_0$
      \State \textbf{Output:} $(provable, trace)$

      \If{timeout or $d > d_{max}$ or $\tau \in V$} \Return $(\bot, \emptyset)$ \EndIf

      \State $V \gets V \cup \{\tau\}$

      \If{$\exists \pi \in \Pi: \tau \equiv \pi$} \Return $(\top, \{\pi\})$ \EndIf

      \State $\mathcal{P} \gets \Call{FindPaths}{\tau, \Pi}$ \Comment{Get inference paths}
      \State Sort $\mathcal{P}$ by $(|\mathcal{I}_p|, \text{priority}(r_p))$ \Comment{$\mathcal{I}_p$: intermediates, $r_p$: rule}

      \For{each path $p \in \mathcal{P}$}
          \State $\Pi_p \gets \emptyset$, $provable \gets \top$
          \For{each intermediate $\iota \in \mathcal{I}_p$}
              \State $(v, \Pi_\iota) \gets \Call{IsProvable}{\iota, \Pi, V', d+1, t_0}$ \Comment{$V' = V$}
              \If{$\neg v$} $provable \gets \bot$; \textbf{break} \EndIf
              \State $\Pi_p \gets \Pi_p \cup \Pi_\iota$
          \EndFor
          \If{$provable$}
              \State $V \gets V \setminus \{\tau\}$
              \State \Return $(\top, \Pi_p)$
          \EndIf
      \EndFor

      \State $V \gets V \setminus \{\tau\}$
      \State \Return $(\bot, \emptyset)$
  \EndFunction
  \end{algorithmic}
  \end{algorithm}

\begin{algorithm}[t]
  \small
  \caption{Backward Chaining Proof Search}
  \begin{algorithmic}[1]
  \Function{IsProvable}{$\tau, \Pi, V, d$}
      \If{$d > d_{max}$ or $\tau \in V$} \Return $\bot$ \EndIf
      \If{$\exists \pi \in \Pi: \tau \equiv \pi$} \Return $\top$ \EndIf
      \State $V \gets V \cup \{\tau\}$
      \For{path $(r, \mathcal{I}) \in \Call{FindPaths}{\tau, \Pi}$}
          \If{$\forall \iota \in \mathcal{I}: \Call{IsProvable}{\iota, \Pi, V, d{+}1}$}
              \State \Return $\top$
          \EndIf
      \EndFor
      \State \Return $\bot$
  \EndFunction
  \end{algorithmic}
  \end{algorithm}

  This function implements a backward chaining algorithm that:
  1. Checks if a target conclusion can be derived from given premises
  2. Uses memoization (visited set) to prevent circular reasoning
  3. Enforces timeout and depth limits to prevent infinite loops
  4. First checks if the target is directly in premises (base case)
  5. Then explores multiple reasoning paths using inference rules (MP, CE, CI, etc.)
  6. Recursively proves intermediate steps needed for each path
  7. Returns the first successful proof path found, with optional trace information.

\paragraph{\texttt{get\_equivalent\_depth}}
\begin{algorithm}[h]
  \caption{ - Get Equivalent Depth}
  \label{alg:get-equivalent-depth}
  \footnotesize
  \begin{algorithmic}
  \Function{GetEquivalentDepth}{node, logTree}
      \State n \(\gets\) \Call{FindLogNodeByOutput}{node.original}
      \If{n \(\neq\) null} \Return n.depth \(-\) 1 \EndIf
      \State \((ok, tr) \gets \Call{IsProvable}{node, \{s \in stmtList \mid s.type=\text{``premise''}\}}\)
      \If{\(\neg\) ok} \Return \(-1\) \EndIf
      \State used \(\gets \{p.original \mid p \in tr.usedPremises\}\)
      \State minD \(\gets\) logScale;\; exact \(\gets\) false
      \ForAll{n \textbf{in} logTree}
          \If{n.depth \(<\) minD \(\wedge\) (n.req \(\subseteq\) used \(\vee\) \Call{Pred}{n.output} \(=\) node.output)}
              \State minD \(\gets\) n.depth;\; exact \(\gets\) (n.req \(=\) used)
          \EndIf
      \EndFor
      \State \Return \textbf{if} minD \(=\) logScale \textbf{then} 0 \textbf{elif} exact \textbf{then} minD\(-\)1 \textbf{else} \Call{Max}{minD\(-\)2, 0}
  \EndFunction
  \end{algorithmic}
  \end{algorithm}

This function computes an ``equivalent depth'' for a logical node by:
1. Checking if it already exists in the LoG tree (returns depth-1)
2. Attempting to prove it from premises and tracking which premises are used
3. Finding the shallowest matching node in the LoG tree via two strategies: Premise coverage: nodes whose required premises are covered by the proof; Output predicate matching: nodes with identical output predicates
4. Returning adjusted depth based on match quality (exact match: depth-1, partial match: depth-2).

\paragraph{planning and reflection's context window}
\label{para:plan-ref-window}
The effective context window for planning nodes comprises the 5 trailing sentences relative to the planning node position, inclusive of the current sentence under consideration.

Similarly, the impact window for reflection sentences extends over a span of 5 sentences when computing reflection-associated metrics, such as interval gain.

\section{Experiment Supplement}
\label{app:experiment_supplement}
\subsection{Details for Models}
\label{app:model_details}
Table~\ref{tab:model_training_methods} shows thinking Mode, Model Source, training method and training data for each models.
\begin{table*}[t]
    \centering
    \small
    \begin{tabular}{l c c l l}
        \toprule
        Model & Thinking Mode & Source & Training Method & Data \\
        \midrule
        \textbf{\textit{Qwen Series}} \\
        QwQ-32B & Long & Open & SFT + RL & reason + general \\
        Qwen3-235B & Long / Short & Open & SFT + RL & reason + general \\
        Qwen3-235B-Instruct & Short & Open & - & reason + general \\
        Qwen3-235B-thinking & Long & Open & - & reason + general \\
        Qwen3-30B & Long / Short & Open & SFT & reason + general \\
        Qwen3-30B-Instruct & Short & Open & - & reason + general \\
        Qwen3-30B-thinking & Long & Open & - & reason + general \\
        Qwen3-4B/8B/14B & Long / Short & Open & SFT & reason + general \\
        Qwen3-32B & Long / Short & Open & SFT + RL & reason + general \\
        Qwen2.5-32B-Instruct & Short & Open & SFT & general \\
        \midrule
        \textbf{\textit{Claude Series}} \\
        Claude-Opus-4.5 & Long/Short & Closed & - & - \\
        Claude-Sonnet-4.5 & Long/Short & Closed & - & - \\
        \midrule
        \textbf{\textit{DeepSeek Series}} \\
        DeepSeek-R1(671B) & Long & Open & SFT + RL & reason + general \\
        DS-R1-Qwen-7B & Long & Open & SFT & reason + general \\
        DS-R1-Qwen-32B & Long & Open & SFT & reason + general \\
        \bottomrule
    \end{tabular}
    \caption{Model training methods classification, detailing Thinking Mode, Source (Open/Closed), Training Method (SFT, RL), and Data type for models across Qwen, Claude, and DeepSeek series.}
    \label{tab:model_training_methods}
\end{table*}

\subsection{Evaluation metrics}
\label{app:full_metrics}
In this subsection, we enumerate all metrics implemented in our codebase. The metrics are organized into a two-tier structure: (1) a high-level taxonomy grouping metrics into \textit{Core} and \textit{Diagnostic} categories (Table~\ref{tab:metrics_taxonomy}), which guide process-level evaluation; and (2) a comprehensive list of base and derived metric implementations at sentence and node levels (Table~\ref{tab:metrics}).

\begin{table*}[t]
\centering
\small
\begin{tabular}{c|l|l|l}
\toprule
\textbf{Category} & \multicolumn{1}{c|}{\textbf{Metric}} & \multicolumn{1}{c|}{\textbf{Code Metric}} & \multicolumn{1}{c}{\textbf{Meaning}} \\
\midrule
\multirow{4}{*}{\textbf{Core}} & \emph{Logical Depth} ($S_\textit{ld}$) & Max Depth & Average maximum correct logical depth achieved \\
\cmidrule{2-4}
& \multirow{3}{*}{\emph{Cost} ($S_\textit{cost}$)} & Reflection Count & \# of reflection steps during reasoning \\
& & Planning Count & \# of planning nodes during reasoning \\
& & Verbosity & square root of \# of tokens generated during reasoning \\

\midrule
\multirow{9}{*}{\textbf{Diagnostic}} & \emph{Exploration} & Explored Node Count & \# of actual nodes provable from given premises \\
\cmidrule{2-4}
& \multirow{2}{*}{\emph{Efficiency}} & Step Efficiency & Effective logical depth advancement per token \\
& & Effective Span & Relative position of last novel and correct step \\
\cmidrule{2-4}
& \multirow{2}{*}{\emph{Coherence}} & Reflection Efficiency & Depth gain induced by reflection in a given window \\
& & Valid Planning & Proportion of  \textit{valid }planning steps executed \\
\cmidrule{2-4}
& \multirow{2}{*}{\emph{Redundancy}} & Sentence Duplication & Proportion of duplicated sentences during reasoning \\
& & Node Duplication Ratio & Repeated visits to identical logical nodes \\

\bottomrule
\end{tabular}
\caption{Taxonomy of reasoning behavior metrics. Core metrics define the primary evaluation axes (Logical Depth, Cost), while diagnostic metrics capture process-level characteristics (Exploration, Efficiency, Coherence, Redundancy) underlying reasoning performance. Within each group, all metrics are first normalized to ensure comparability across metrics, and are then aggregated—by weighted averaging—into a single group-level score. For verbosity, we apply a square root transformation prior to normalization to mitigate the high variance in raw token counts.}
\label{tab:metrics_taxonomy}
\end{table*}

\begin{table*}[t]
\centering
\footnotesize
\begin{tabular}{c|c|l|l|l}
\toprule
\textbf{Category} & \textbf{Level} & \textbf{Type} & \textbf{Metric} & \textbf{Meaning} \\
\midrule
\multirow{13}{*}{\textbf{Base}} & \multirow{6}{*}{Node} & \multirow{3}{*}{Actual node} & Count & Number of premise/derived/hallucination nodes \\
& & & Correctness & Whether actual node is correct with given premise \\
& & & Depth & Log node depth or calculated equivalent depth \\
\cmidrule(lr){3-5}
& & \multirow{3}{*}{Planning node} & Count & Number of planning nodes \& unique \\
& & & Correctness & Statement can be proved with given premises \\
& & & Effectiveness/valid & New actual node achieved from planning in given window \\
\cmidrule(lr){2-5}
& \multirow{7}{*}{Sentence} & \multirow{3}{*}{Verbosity} & Sentence count & Number of sentences during reasoning \\
& & & Token count & Number of tokens during reasoning \\
& & & Node count & Number of nodes during reasoning \\
\cmidrule(lr){3-5}
& & Reflection & Sentence count & Number of reflection sentences \\
\cmidrule(lr){3-5}
& & \multirow{3}{*}{Reflection effect} & Has new node & Count if reflection window creates new node \\
& & & Has deeper node & Count if reflection window creates deeper node \\
& & & Has new hallucination & Count if reflection window creates new hallucination node \\
\midrule
\multirow{18}{*}{\textbf{Derived}} & \multirow{8}{*}{Node} & All node & Node Duplication Ratio & Repeated visits to identical logical nodes \\
\cmidrule(lr){3-5}
& & \multirow{8}{*}{Actual node} & Exploration precision & Proportion of explored nodes on minimum log graph \\
& & & Reasoning accuracy & Proportion of correct nodes on minimum log graph \\
& & & Premise coverage & Proportion of premise used \\
& & & Depth coverage & Proportion of depth covered during reasoning \\
& & & Depth & Average maximum correct logical depth achieved \\
& & & Incorrect ratio & Proportion of hallucination actual nodes \\
& & & Interval depth & Length of complete reasoning subtree interval \\
\cmidrule(lr){2-5}
& \multirow{8}{*}{Sentence} & \multirow{4}{*}{Efficiency} & First correct step & First sentence containing correct nodes \\
& & & Step Efficiency & Effective logical depth advancement per expenditure \\
& & & Node Efficiency & Average correct nodes per sentence \\
& & & Reflection Efficiency & Depth gain induced by reflection in given window \\
\cmidrule(lr){3-5}
& & \multirow{3}{*}{Spans} & Effective Span & Relative position of last novel and correct step \\
& & & Forward reasoning span & Relative position of last novel and deepest step \\
& & & Reasoning span & Relative position of last novel step \\
\cmidrule(lr){3-5}
& & Duplication & Sentence duplication ratio & Proportion of duplicated sentences during reasoning \\
\bottomrule
\end{tabular}
\caption{Full reasoning evaluation metrics, categorized into Base (node/sentence-level counts and correctness) and Derived (aggregated performance indicators like exploration, efficiency, span, and duplication ratios) for comprehensive behavioral analysis.}
\label{tab:metrics}
\end{table*}

\paragraph{node level metrics.}
For node definition, refer to Table~\ref{tab:node_def}. We evaluate actual nodes by their count (premise/derived/hallucination), correctness, and depth. Planning nodes are assessed by count, correctness, and effectiveness (whether they generate new actual nodes within a given window). Derived metrics include exploration precision, reasoning accuracy, premise and depth coverage, node duplication ratio, and incorrect ratio.

\paragraph{sentence level metrics.} 
We track basic verbosity (sentence count, token count, node count) and distinguish reflection sentences by their count and ratio. Efficiency metrics include first correct step, step efficiency (depth advancement per expenditure), node efficiency (correct nodes per sentence), and reflection efficiency. We measure reasoning spans (effective, forward, and overall) based on the relative position of the last novel step. For reflection sentences specifically, we track whether their context windows produce new nodes, deeper nodes, or hallucinations. Finally, we compute sentence duplication ratio to identify repeated reasoning patterns.

\subsection{Experiment details}
\label{app:exp_setting_detail}
The hyperparameters employed for API invocation comprised: temperature = 0, maximum token allocation of 24,000 for reasoning-enabled models and 8,000 for non-reasoning models, with all other API parameters maintained at their default values.

\noindent The prompt we used for each LoG question is as follows:






\vspace{\baselineskip}
\noindent
\begin{minipage}{\columnwidth}
\begin{mdframed}[backgroundcolor=gray!10]
``Please answer the question based on the given information:

\textbf{Given Information}: \{tmp\_information\}

\textbf{Note}: In this context, `A is B' has the same meaning as `a rabbit is a mammal' --- it means A belongs to category B, not that A equals B.

\textbf{Question}: \{tmp\_question\}

Please reason step by step, show your reasoning process and put your final answer in \textbackslash boxed\{\}.''
\end{mdframed}
\end{minipage}

\subsection{K-means Classification Result}
\label{app:kmeans_classification}
To move beyond isolated metric analysis, we employ a semi-supervised clustering framework to identify generalized behavioral archetypes:

\noindent(1) \textbf{Unsupervised Clustering \& Semantic Mapping:} We first apply K-means clustering ($k=4$) on the normalized 2D feature space defined by Logical Depth and Cost. To interpret the clusters, we define four ``Ideal Archetypes'' corresponding to the quadrants of the plane (Table~\ref{tab:archetype_def}) and utilize the Hungarian Algorithm to map empirical centroids to these semantic labels.

\begin{table}[ht]
    \centering
    \small
    \resizebox{\columnwidth}{!}{
    \begin{tabular}{l|cc|l}
    \toprule
    \textbf{Archetype} & \textbf{LD.} & \textbf{Engage.} & \textbf{Behavioral Trait} \\
    \midrule
    \textbf{Effective Solver} & High & Low & \textit{Efficient \& Correct} \\
    \textbf{Deep Wanderer} & High & High & \textit{Exhaustive \& Correct} \\
    \textbf{Lazy Guesser} & Low & Low & \textit{Direct Failure} \\
    \textbf{Hollow Mimic} & Low & High & \textit{Inefficient Failure} \\
    \bottomrule
    \end{tabular}
    }
    \caption{Definition of Behavioral Archetypes based on the Quadrants of the Logical Depth-Cost Plane. }
    \label{tab:archetype_def}
\end{table}

\noindent(2) \textbf{Boundary-Relative Confidence:} 
To rigorously quantify how ``typical'' a model is of its category, we propose a boundary-aware confidence score ($S_\textit{c}$). Unlike simple centroid distance, this metric considers the geometric decision boundaries (Voronoi partitions). For a model point $P$ assigned to cluster centroid $C_{own}$, the confidence is defined:
\begin{equation}
    S_\textit{c}(P) = \min_{C_{enemy}} \left( \min \left( 1.0, \frac{d(P, B)}{d(C_{own}, B)} \right) \right)
\end{equation}
where $d(\cdot, B)$ denotes the perpendicular distance to the boundary. Intuitively, $S_\textit{c}=1.0$ indicates the model resides deeper in its region than the centroid itself (hyper-typical), while $S_\textit{c} \approx 0$ implies the model lies on the decision boundary.

\begin{figure*}[ht] 
    \centering
    \includegraphics[width=\textwidth]{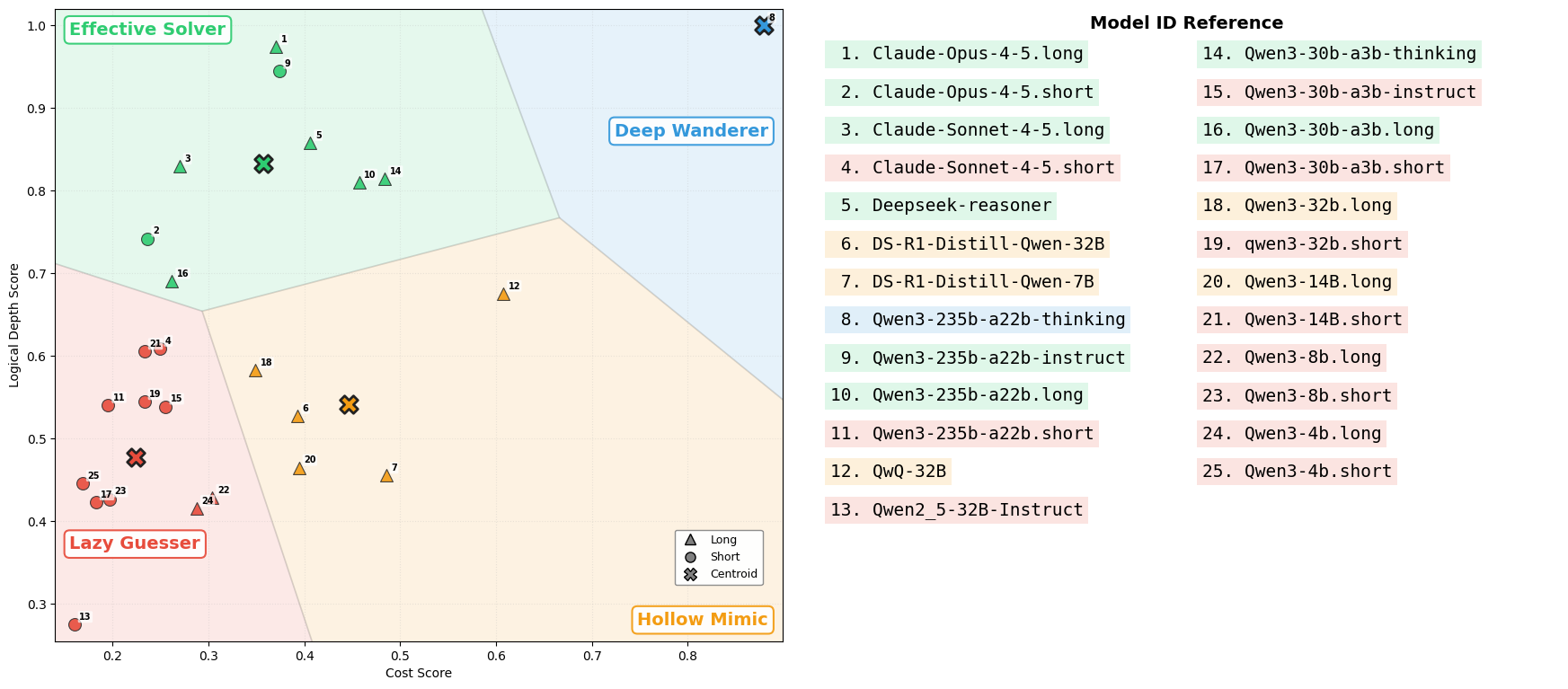}
    \caption{K-means Classification} 
    \label{fig:kmeans_classification} 
\end{figure*}

\subsection{Model Setting Details}
\label{app:model_setting_detail}
Table~\ref{tab:model_training_methods} shows the Training paradigms and datasets of all models evaluated.
Table~\ref{tab:metrics_taxonomy} illustrates the taxonomy of metrics.

\subsection{Experiment Results Details}
\label{app:experiment_details} 
\begin{figure*}[ht]
    \centering
    \includegraphics[width=0.95\linewidth]{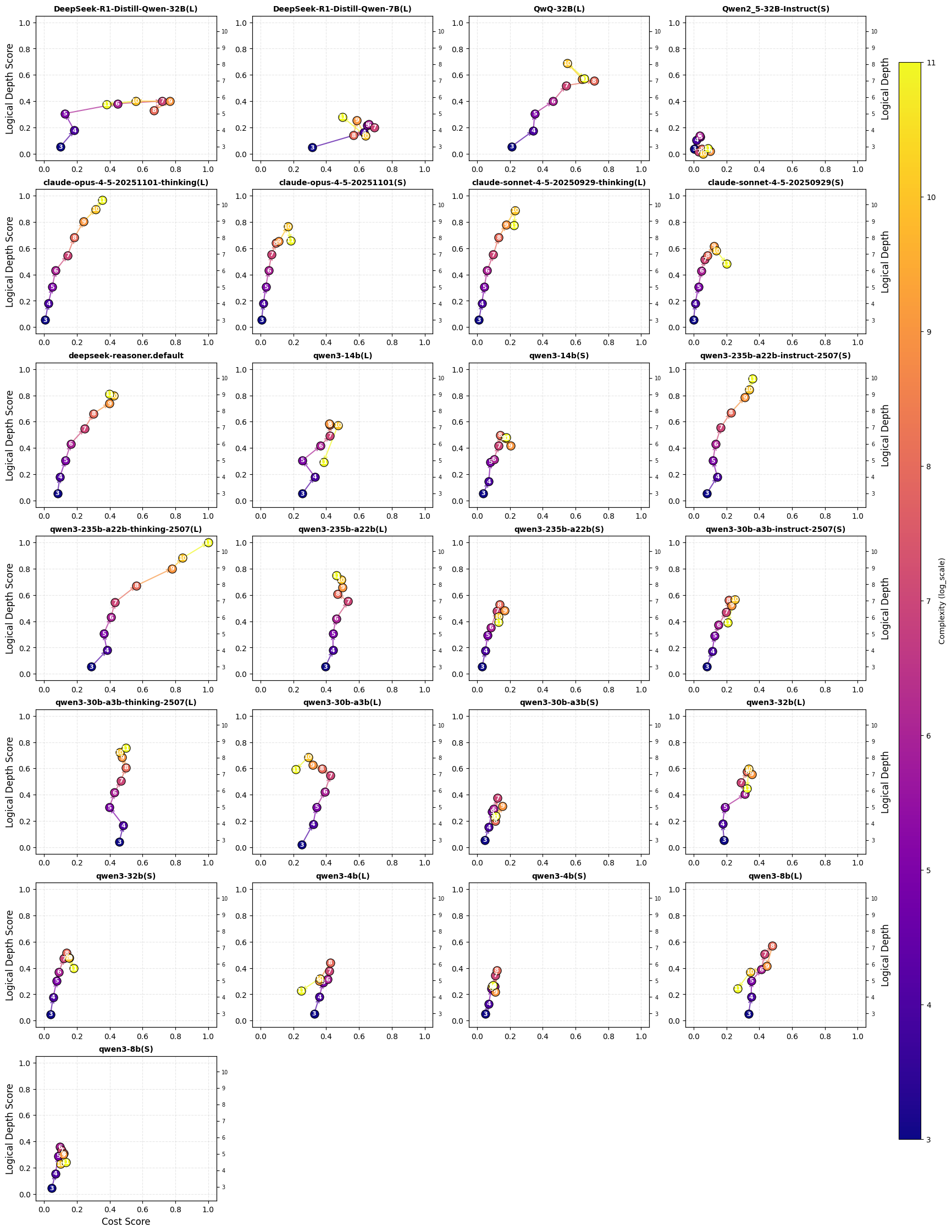}
    \caption{Full set of reasoning trajectories for all 25 evaluated models, plotting Logical Depth Score against Cost Score across increasing complexity (color-coded by depth from log C=3 to 11). Each subplot visualizes a model’s adaptation pattern under varying problem difficulty.}
    \label{fig:trajectory}
\end{figure*}

\begin{figure*}[h]
    \centering
    \includegraphics[width=0.95\textwidth]{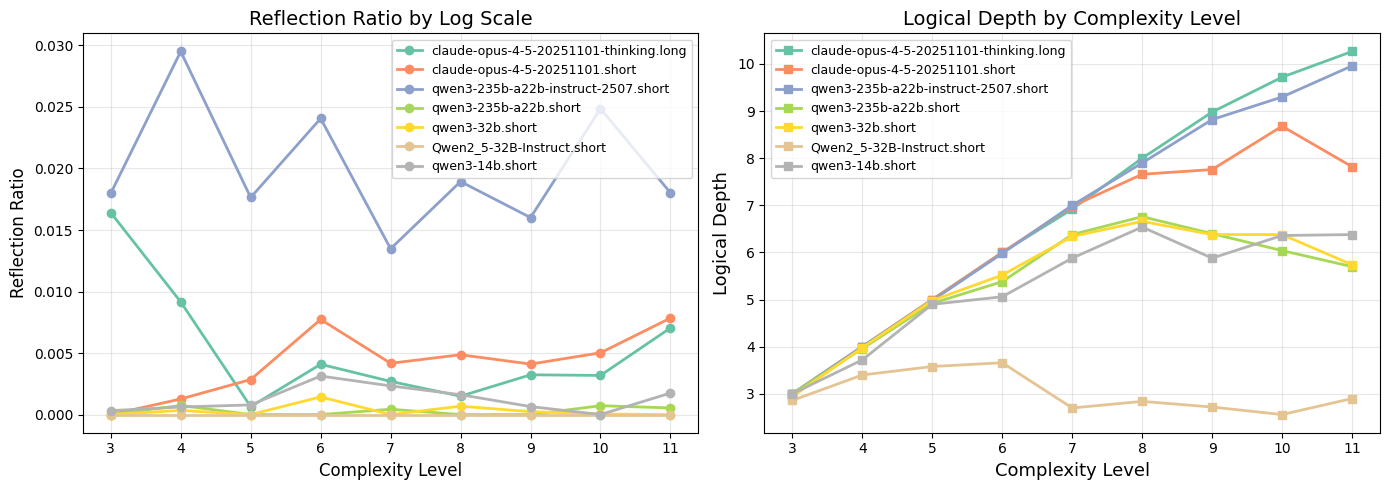}
    \caption{Reflection Ratio and Max Depth by Log Scale. The three models (Qwen3-235B-Instruct, Claude-Opus-4.5.long, Claude-Opus-4.5.short) that maintain an advantage in logical depth in the right figure also demonstrate superior reflection ratios in the left figure. The Qwen3-235B.short model, which shows negligible reflection behavior, significantly lags behind the Qwen3-235B-A33B-Instruct model in logical depth, despite having comparable parameter scales.}
    \label{fig:short_reflection}
\end{figure*}

Table~\ref{tab:model_depth_token_analysis} summarizes the logical depth and the number of tokens used by different models. 
Table~\ref{tab:full_model_classification} is the full version of Table~\ref{tab:model_classification}.
Figure~\ref{fig:trajectory} illustrates the variations in Logical Depth and Cost across all models under different levels of complexity. Figure~\ref{fig:short_reflection} shows the comparison of short models with and without reflection.

\begin{table*}[htbp]
    \centering
    \small
    \begin{tabular}{l l | c c | c c}
        \toprule
        \textbf{Model} & \textbf{Type} & \textbf{Avg Depth} & \textbf{Avg Depth Delta} & \textbf{Avg Token} & \textbf{Avg Token Delta} \\
        \midrule
        QwQ-32B & long & 5.96 & 97.13\% & 4336.67 & 879.18\% \\
        DS-R1-Qwen-32B & long & 5.06 & 67.45\% & 4283.78 & 867.24\% \\
        Qwen2.5-32B-Instruct & short & 3.02 & - & 442.89 & - \\
        \midrule
        Claude-Opus-4.5 & long & 6.88 & 8.76\% & 1398.00 & 108.24\% \\
        & short & 6.32 & - & 671.33 & - \\
        \midrule
        Claude-Sonnet-4.5 & long & 6.68 & 14.20\% & 931.67 & 41.66\% \\
        & short & 5.85 & - & 657.67 & - \\
        \midrule
        Qwen3-235B-thinking & long & 6.88 & 1.61\% & 6850.44 & 295.70\% \\
        Qwen3-235B-Instruct & short & 6.77 & - & 1731.22 & - \\
        \midrule
        Qwen3-235B & long & 6.33 & 17.31\% & 4127.33 & 429.30\% \\
        & short & 5.39 & - & 779.78 & - \\
        \midrule
        Qwen3-32B & long & 5.76 & 6.05\% & 2161.22 & 152.58\% \\
        & short & 5.44 & - & 855.67 & - \\
        \midrule
        Qwen3-30B-thinking & long & 6.28 & 12.63\% & 4582.00 & 242.34\% \\
        Qwen3-30b-Instruct & short & 5.58 & - & 1338.44 & - \\
        \midrule
        Qwen3-30B & long & 6.08 & 36.97\% & 2416.22 & 217.65\% \\
        & short & 4.44 & - & 760.67 & - \\
        \midrule
        Qwen3-14B & long & 5.64 & 6.45\% & 3099.78 & 224.58\% \\
        & short & 5.30 & - & 955.00 & - \\
        \midrule
        Qwen3-8B & long & 5.24 & 14.95\% & 2996.67 & 312.76\% \\
        & short & 4.56 & - & 726.00 & - \\
        \midrule
        Qwen3-4B & long & 4.78 & 6.91\% & 2714.56 & 289.52\% \\
        & short & 4.47 & - & 696.89 & - \\
        \bottomrule
    \end{tabular}
    \caption{Model Depth and Token Analysis, comparing average logical depth, depth delta (relative increase from short to long mode), average token count, and token delta across models and thinking modes.}
    \label{tab:model_depth_token_analysis}
\end{table*}

\begin{table*}[t]
    \centering
    \small
    \setlength{\tabcolsep}{3.5pt} 
    \begin{tabular}{c|l|c|c|cc|cccc|cc}
    \toprule
    \multirow{2}{*}{\#} & \multirow{2}{*}{Model} & \multicolumn{2}{c|}{\textbf{Classification}} & \multicolumn{2}{c|}{\textbf{Core Metrics}} & \multicolumn{4}{c|}{\textbf{Diagnostic Metrics}} & \multicolumn{2}{c}{\textbf{Raw Stats.}} \\
    \cmidrule(lr){3-4} \cmidrule(lr){5-6} \cmidrule(lr){7-10} \cmidrule(lr){11-12}
    & & Category & $S_\textit{c}$ & $S_\textit{ld}$ & $S_\textit{cost}$ & $S_\textit{exp}$ & $S_\textit{eff}$ & $S_\textit{coh}$ & $S_\textit{red}$ & \textbf{Depth} & \textbf{Tok.(k)} \\
    \midrule
    1 & Qwen3-235B-thinking & DeepWanderer & 1.00 & 1.00 & 0.88 & 1.00 & 0.47 & 0.41 & 0.77 & 10.54 & 16.8 \\
    \midrule
    2 & Claude-Sonnet-4.5.long & \multirow{8}{*}{EffectiveSolver} & 0.82 & 0.83 & 0.27 & 0.38 & 0.67 & 0.26 & 0.38 & 8.74 & 1.9 \\
    3 & Qwen3-235B-Instruct & & 0.82 & 0.95 & 0.37 & 0.83 & 0.59 & 0.28 & 0.62 & 9.96 & 3.4 \\
    4 & DeepSeek-R1 & & 0.80 & 0.86 & 0.41 & 0.34 & 0.59 & 0.58 & 0.48 & 9.04 & 3.7 \\
    5 & Claude-Opus-4.5.long & & 0.80 & 0.97 & 0.37 & 0.47 & 0.60 & 0.42 & 0.28 & 10.27 & 3.5 \\
    6 & Qwen3-235B.long & & 0.67 & 0.81 & 0.46 & 0.29 & 0.57 & 0.31 & 0.55 & 8.54 & 4.1 \\
    7 & Qwen3-30B-thinking & & 0.58 & 0.81 & 0.48 & 0.06 & 0.52 & 0.50 & 0.53 & 8.58 & 5.3 \\
    8 & Claude-Opus-4.5.short & & 0.33 & 0.74 & 0.24 & 0.62 & 0.70 & 0.47 & 0.37 & 7.82 & 1.4 \\
    9 & Qwen3-30B.long & & 0.12 & 0.69 & 0.26 & 0.04 & 0.64 & 0.59 & 0.39 & 7.28 & 1.5 \\
    \midrule
    10 & DS-R1-Qwen-7B & \multirow{5}{*}{HollowMimic} & 1.00 & 0.49 & 0.46 & 0.01 & 0.47 & 0.35 & 0.62 & 4.80 & 6.0 \\
    11 & DS-R1-Qwen-32B & & 0.53 & 0.53 & 0.39 & 0.12 & 0.50 & 0.76 & 0.59 & 5.56 & 3.4 \\
    12 & Qwen3-14B.long & & 0.39 & 0.47 & 0.39 & 0.09 & 0.54 & 0.34 & 0.57 & 4.90 & 3.4 \\
    13 & QwQ-32B & & 0.34 & 0.68 & 0.61 & 0.14 & 0.48 & 0.32 & 0.62 & 7.12 & 5.7 \\
    14 & Qwen3-32B.long & & 0.29 & 0.58 & 0.35 & 0.24 & 0.56 & 0.33 & 0.52 & 6.14 & 2.7 \\
    \midrule
    15 & Qwen2.5-32B-Inst & \multirow{11}{*}{LazyGuesser} & 1.00 & 0.28 & 0.16 & 0.03 & 0.55 & 0.07 & 0.59 & 2.90 & 0.7 \\
    16 & Qwen3-30B.short & & 1.00 & 0.42 & 0.18 & 0.06 & 0.63 & 0.07 & 0.51 & 4.46 & 0.8 \\
    17 & Qwen3-8B.short & & 1.00 & 0.43 & 0.20 & 0.05 & 0.63 & 0.17 & 0.49 & 4.50 & 1.0 \\
    18 & Qwen3-4B.short & & 1.00 & 0.45 & 0.17 & 0.02 & 0.67 & 0.11 & 0.54 & 4.70 & 0.7 \\
    19 & Qwen3-235B.short & & 0.74 & 0.54 & 0.20 & 0.12 & 0.70 & 0.43 & 0.53 & 5.70 & 1.0 \\
    20 & Qwen3-32B.short & & 0.65 & 0.55 & 0.23 & 0.18 & 0.64 & 0.41 & 0.53 & 5.74 & 1.4 \\
    21 & Qwen3-4B.long & & 0.62 & 0.42 & 0.29 & 0.03 & 0.55 & 0.39 & 0.48 & 4.38 & 1.7 \\
    22 & Qwen3-30B-Instruct & & 0.59 & 0.54 & 0.26 & 0.27 & 0.58 & 0.47 & 0.45 & 5.68 & 1.6 \\
    23 & Qwen3-8B.long & & 0.45 & 0.43 & 0.30 & 0.07 & 0.55 & 0.40 & 0.44 & 4.52 & 1.9 \\
    24 & Qwen3-14B.short & & 0.35 & 0.23 & 0.61 & 0.08 & 0.68 & 0.23 & 0.55 & 6.38 & 1.4 \\
    25 & Claude-Sonnet-4.5.short & & 0.30 & 0.61 & 0.25 & 0.34 & 0.61 & 0.42 & 0.40 & 6.42 & 1.6 \\
    \midrule
    \multicolumn{2}{c|}{\multirow{4}{*}{\textit{Category Avg (weighted)}}} & DeepWanderer & -- & \textbf{1.00} & \textbf{0.88} & \textbf{1.00} & 0.47 & 0.41 & \textbf{0.77} & 10.54 & 16.8 \\
    \multicolumn{2}{c|}{} & EffectiveSolver & -- & 0.86 & 0.37 & 0.42 & 0.60 & 0.40 & 0.46 & 9.40 & 3.1 \\
    \multicolumn{2}{c|}{} & HollowMimic & -- & 0.52 & 0.45 & 0.09 & 0.50 & \textbf{0.42} & 0.60 & 5.70 & 4.2 \\
    \multicolumn{2}{c|}{} & LazyGuesser & -- & 0.45 & 0.21 & 0.09 & \textbf{0.62} & 0.25 & 0.51 & 5.03 & 1.2 \\
    \bottomrule
    \end{tabular}
    \caption{Full model classification results. The \textbf{Raw Stats.} columns (rightmost) display absolute Average Logical Depth and Token Count (in thousands) for reference. Note that the category averages are weighted by confidence scores ($S_\textit{c}$).}
    \label{tab:full_model_classification}
\end{table*}

\section{Analysis Supplement}
\label{app:analysis_supplement}
\subsection{analysis plots}
\label{app:analysis_plots}

\begin{figure*}[htp]
    \centering
    \begin{subfigure}[b]{0.32\textwidth}
        \centering
        \includegraphics[height=4cm]{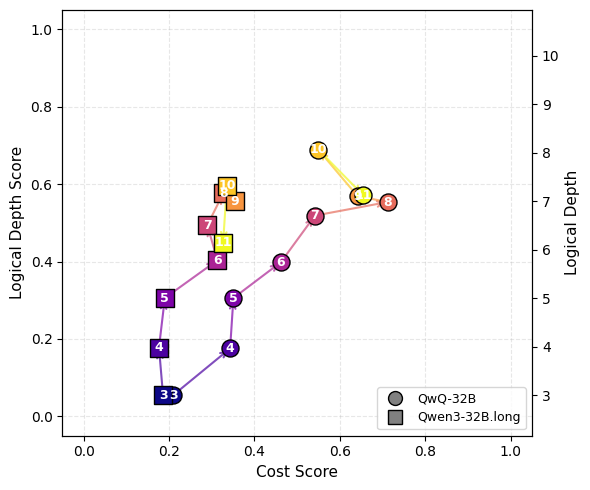}
        \caption{32B Seperate vs Mixed}
        \label{fig:qwen3_32b_vs_qwq}
    \end{subfigure}
    \hfill
    \begin{subfigure}[b]{0.32\textwidth}
        \centering
        \includegraphics[height=4cm]{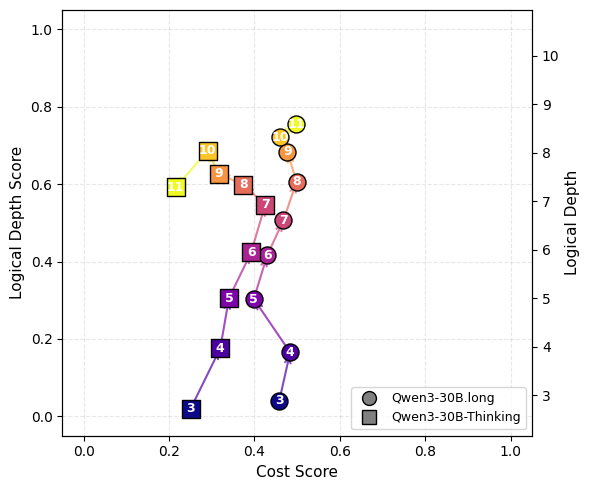}
        \caption{30B Seperate vs Mixed}
        \label{fig:qwen3_30b_a3b}
    \end{subfigure}
    \hfill
    \begin{subfigure}[b]{0.32\textwidth}
        \centering
        \includegraphics[height=4cm]{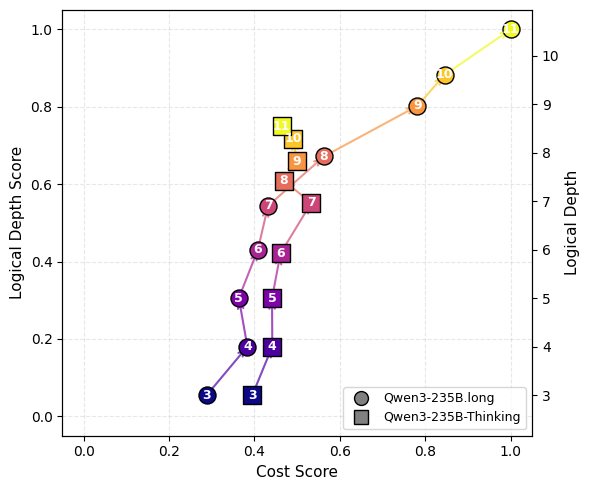}
        \caption{235B Seperate vs Mixed}
        \label{fig:qwen3_235b_a22b}
    \end{subfigure}

    \caption{Comparative Analysis of Reasoning Models. Performance trajectories of (a) 32B, (b) 30B, and (c) 235B model variants under separate vs. mixed configurations, plotting Logical Depth against Cost across increasing complexity. It can be observed that, in these three settings, the independently trained thinking models (QwQ-32B, Qwen3-30B-thinking, Qwen3-235B-thinking) experience saturation or collapse later than their counterparts trained with long/short mixed methods. This reflects the disruptive effect of mixed training on high-consumption strategies.}
    \label{fig:seperate_vs_mixed}
\end{figure*}

\begin{figure*}[ht]
    \centering
    \includegraphics[width=0.95\textwidth]{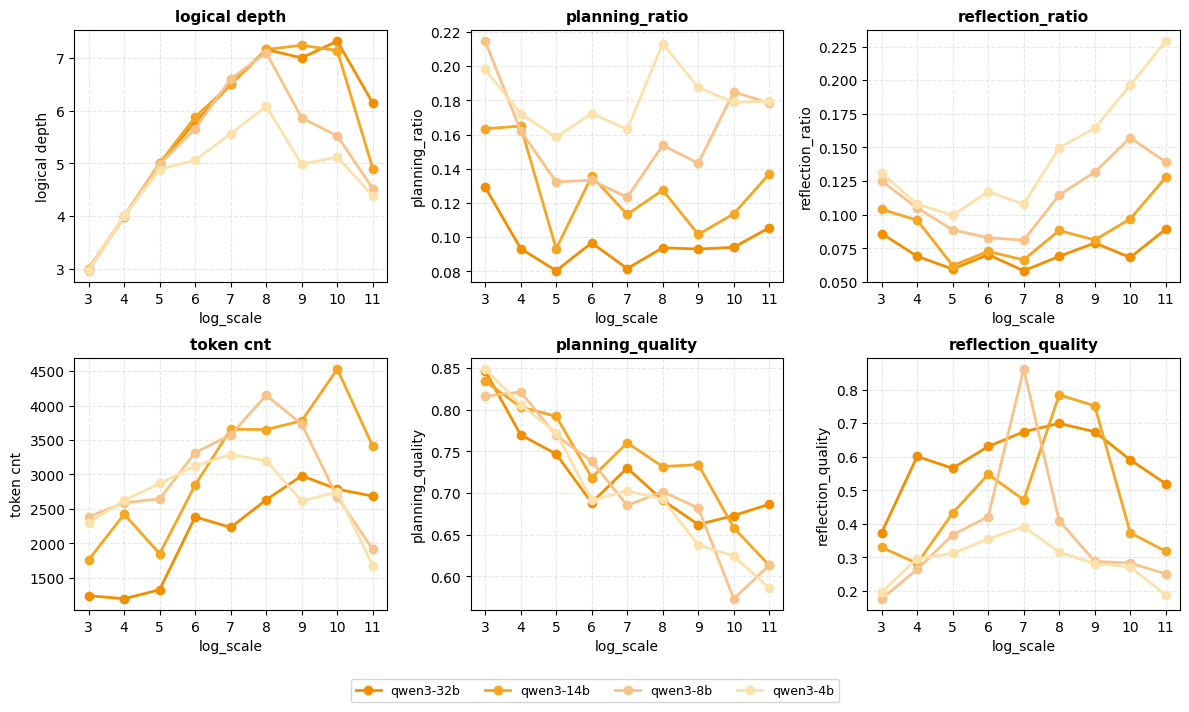}
    \caption{Qwen3-32B vs 14B vs 8B vs 4B. Six subplots compare key reasoning behaviors (max depth, planning/reflection ratio and quality, token count) across model sizes as problem complexity (log\_scale) increases. Smaller distilled models exhibit more sophisticated behaviors, even surpassing the teacher model; however, they fail to emulate behavioral efficiency and cannot translate these sophisticated behaviors into deeper reasoning. Only the 14B model shows a high degree of alignment in both behavior and capabilities with the 32B teacher model. This indicates that the effectiveness of the model's reasoning in expanding tokens is intrinsically constrained by its capabilities.}
    \label{fig:distill_comparison}
\end{figure*}

We present additional empirical insights through two key comparative visualizations. Figure~\ref{fig:seperate_vs_mixed} illustrates the performance divergence between separately trained and mixed training configurations across 32B, 30B, and 235B reasoning models, highlighting the detrimental impact of mixed training on sustaining deep logical Cost. Complementarily, Figure~\ref{fig:distill_comparison} examines the behavioral fidelity of distilled smaller models (4B–14B) relative to the Qwen3-32B teacher, revealing that while smaller models can mimic sophisticated reasoning patterns, only the 14B variant successfully aligns both behaviorally and capability-wise with the teacher, underscoring the intrinsic limitations of token-efficient reasoning in under-capacitated models.

\end{document}